\documentclass{article}

\usepackage[nonatbib,preprint]{neurips_data_2022}

\usepackage[utf8]{inputenc} 
\usepackage[T1]{fontenc}    
\usepackage{hyperref}       
\usepackage{url}            
\usepackage{booktabs}       
\usepackage{amsfonts}       
\usepackage{nicefrac}       
\usepackage{microtype}      
\usepackage{xcolor}         

\usepackage{graphicx} 
\usepackage{float} 
\usepackage{subfigure} 
\usepackage{color}
\usepackage{threeparttable}
\usepackage{booktabs}
\usepackage{multirow}
\usepackage{amsmath,mathtools,amssymb}
\usepackage{subcaption} 
\usepackage{makecell}

\newcommand{\ourproj}[1]{\emph{Droplet3D}}
\newcommand{\ourdataset}[1]{\emph{Droplet3D-4M}}

\usepackage{tabularx}

\title{Droplet3D: Commonsense Priors from Videos Facilitate 3D Generation}

\author{
Xiaochuan Li$^{1,}$\footnotemark[1] ~,
Guoguang Du$^{1,}$\footnotemark[1] ~,
Runze Zhang$^{1,}$\footnotemark[1] ~,
Liang Jin$^{1,}$\footnotemark[1] ~,
Qi Jia$^{1,}$\footnotemark[1] ~,
Lihua Lu$^{1,}$\footnotemark[1] ~
\\
\textbf{Zhenhua Guo$^{1}$},
\textbf{Yaqian Zhao$^{1}$},
\textbf{Haiyang Liu},
\textbf{Tianqi Wang},
\textbf{Changsheng Li},
\textbf{Xiaoli Gong$^{2}$}
\\
\textbf{Rengang Li$^{3,1,}$\footnotemark[2] ~,}
\textbf{Baoyu Fan$^{2,1,}$\footnotemark[2] ~}
\\
  $^1$ IEIT System Co., Ltd. ~~
  $^2$ Nankai University ~~
  $^3$ Tsinghua University
  \\
  \url{https://dropletx.github.io}
}

\begin{document}

\maketitle
\renewcommand{\thefootnote}{\fnsymbol{footnote}}
\footnotetext[1]{Equal contribution.}
\footnotetext[2]{Corresponding author.}


\begin{figure}[htbp]
\centering
\includegraphics[width=1.0\linewidth]{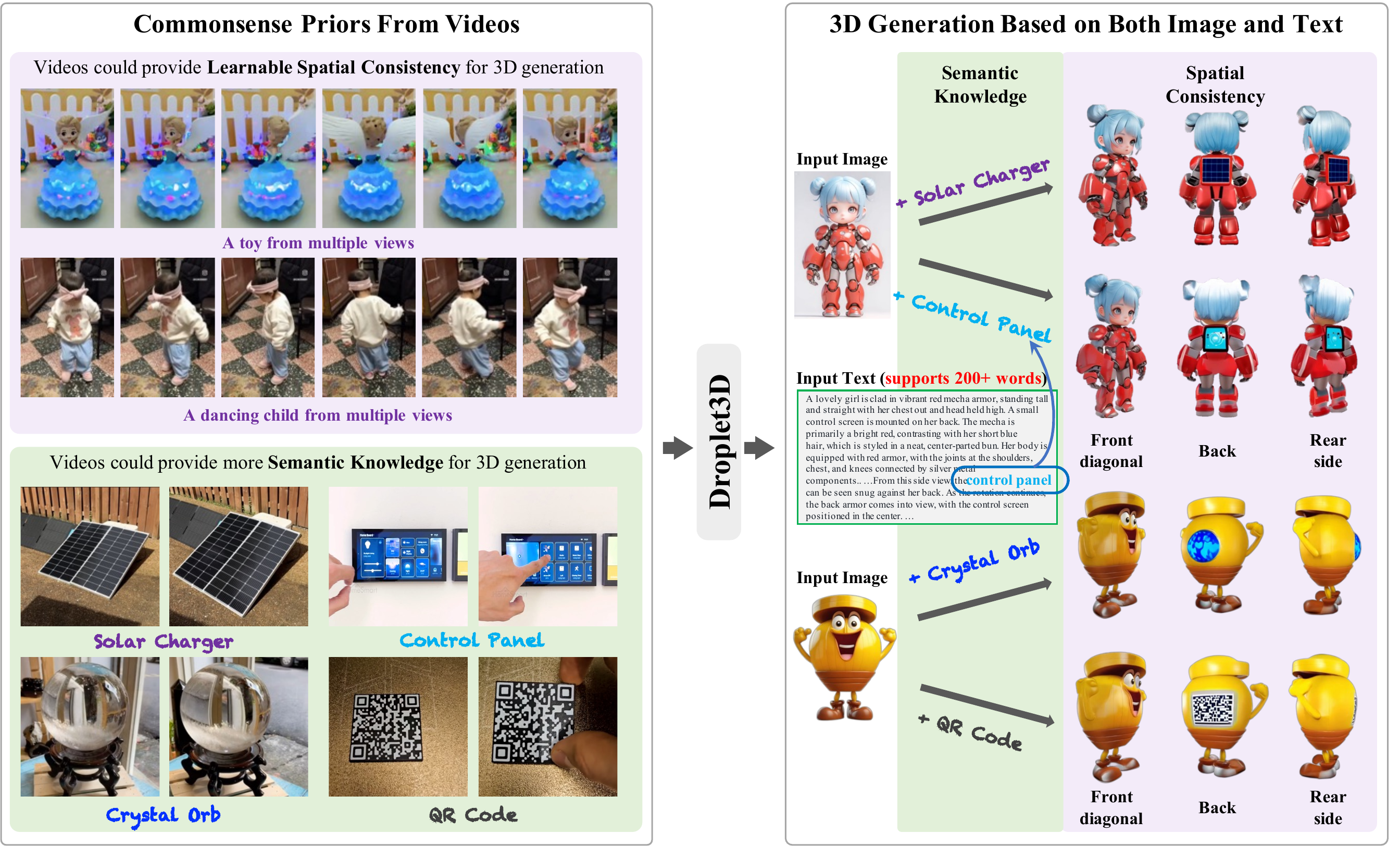}
\caption{\textbf{\ourproj{} achieves creative 3D content generation based on both image and text input.} Commonsense priors including spatial consistency and semantic knowledge facilitate the 3D generation abilities of our method.}
\label{fig:Title}
\end{figure}

\begin{abstract}



Scaling laws have validated the success and promise of large-data-trained models in creative generation across text, image, and video domains. 
However, this paradigm faces data scarcity in the 3D domain, as there is far less of it available on the internet compared to the aforementioned modalities.
Fortunately, there exist adequate videos that inherently contain commonsense priors, offering an alternative supervisory signal to mitigate the generalization bottleneck caused by limited native 3D data. 
On the one hand, videos capturing multiple views of an object or scene provide a spatial consistency prior for 3D generation. 
On the other hand, the rich semantic information contained within the videos enables the generated content to be more faithful to the text prompts and semantically plausible.
This paper explores how to apply the video modality in 3D asset generation, spanning datasets to models. 
We introduce \ourdataset{}, the first large-scale video dataset with multi-view level annotations, and train \ourproj{}, a generative model supporting both image and dense text input. 
Extensive experiments validate the effectiveness of our approach, demonstrating its ability to produce spatially consistent and semantically plausible content. 
Moreover, in contrast to the prevailing 3D solutions, our approach exhibits the potential for extension to scene-level applications.
This indicates that the commonsense priors from the videos significantly facilitate 3D creation. 
We have open-sourced all resources including the dataset, code, technical framework, and model weights: \url{https://dropletx.github.io/}.

\end{abstract}

\section{Introduction}

Thanks to the increasing availability of massive online data, large-scale pre-trained models have demonstrated such generative potential in multi-modalities, such as text\cite{2303.08774,radford2018improving}, image\cite{rombach2022high,li2024image}, and video\cite{openai-sora,zhang2025dropletvideo}.
However, the scaling of 3D generative models faces a significant bottleneck due to data scarcity, as the collection of 3D data is considerably more challenging than for other, more readily available modalities.
For example, the largest open-source 3D dataset, Objaverse-XL \cite{deitke2023objaverse,deitke2023objaverse2,lin2025objaversepp}, contains merely 10 million samples, orders of magnitude smaller than the image–text paired datasets\cite{schuhmann2022laion} which usually consists of billions of samples.
This presents two main challenges.
First, the limited coverage of existing 3D datasets hinders the comprehensive capture of the real world, leading to a scarcity of learnable spatial consistency information that is constrained by data distribution biases. 
Second, compared to modalities such as text and images, 3D generative models learn less semantic knowledge, restricting their capacity to generate diverse content. 
Consequently, while the scaling law has been extensively validated in other modalities, the progress of this paradigm in the 3D domain remains comparatively slow.

Fortunately, the abundance of online video data offers a promising solution to mitigate 3D data scarcity\cite{wang2024nova,kang2025consistent,wu2025genfusion}.
First, vast video resources inherently contain multi-view spatial consistency information, as illustrated by the ``toy'' and ``dancing child'' examples in Fig. \ref{fig:Title} where objects are captured from multiple views. 
From this viewpoint, such videos can be regarded as a specific representation of 3D data. 
Introducing these videos significantly enhances the foundational capability of 3D generative models to maintain cross-view consistency.
Second, the scale advantage of video datasets enables broader semantic knowledge than 3D data. 
As demonstrated by the ``QR code'' case in Fig. \ref{fig:Title}, by learning from video samples, a 3D model can comprehend semantic concepts not present in its 3D training corpus and successfully generate objects that conform to textual descriptions. 
This indicates that the diverse semantic knowledge encapsulated in videos provides 3D models with critical generalizable commonsense priors.
Motivated by this, we hope to study a video-driven paradigm for 3D generation, which creates 3D content based on an image and detailed text prompts.
A promising approach involves fine-tuning pretrained video generation models with 3D data, enabling the adapted 3D generation models to inherit beneficial commonsense priors from the video domain. 

A promising paradigm involves fine-tuning a pre-trained video generation model using 3D data, enabling the resulting 3D model to inherit the rich commonsense priors from its video counterpart, thereby enhancing its generative capabilities. 
Although existing work\cite{chen2024v3d,voleti2024sv3d} recognized this potential, their specific fine-tuning datasets, base models, and strategies exhibited limitations. 
Specifically, these approaches fail to adequately leverage text supervision and lack explicit mechanisms to preserve the capabilities transferred from the video backbone.

In light of this, our research on this issue encompasses from dataset construction to model development.
We proposed a large-scale 3D dataset, \ourdataset{}.
It comprises approximately 4 million 3D models, each accompanied by a 360° orbital rendering video and a detailed multi-view level text annotation. 
Each video consists of 85 frames captured along a circular camera path around the object, which covers a complete 360-degree view. 
On the text side, each video is paired with a dense natural language caption averaging about 260 words in length. 
This caption meticulously characterizes the object's overall appearance and elaborates on the unique details of different views, such as front, side, or back views, from sequential viewpoints as the camera orbits. 
This dense multi-view-level text annotation provides rich supervisory signals for the model to understand and generate each viewpoint, which benefits 3D generative models developing.

Furthermore, we introduce a novel 3D generative model, \ourproj{}, which supports both image and dense text input, allowing our model to generate content that meets user needs with greater granularity. 
It is fine-tuned from DropletVideo\cite{zhang2025dropletvideo}, a video diffusion model capable of modeling integral spatio-temporal consistency. 
\ourproj{} is capable of generating a sequence of 85 dense multi-view images capturing a full 360-degree orbit of the object. 
In addition, to adapt to the input requirements of \ourproj{}, we introduce a text rewriting module and an initial image canonical view alignment module to preprocess the user's prompt, thereby aligning it with the distribution of \ourdataset{}.
Finally, based on the generated orbital views, we employed downstream applications using both 3D Gaussian splatting and textured mesh reconstruction algorithms, surpassing the fixed output format commonly used in current prevailing 3D generation solutions\cite{zhao2025hunyuan3d20,lai2025hunyuan3d25}. 

Comprehensive qualitative and quantitative experiments demonstrate the superiority of our proposed approach for 3D content generation tasks based on images and text. 
Extensive visualization experiments show various interesting capabilities of \ourproj{}, demonstrating the potential of video-driven 3D generation methods for creative generation.
Notably, in contrast to the prevailing object-level 3D solutions\cite{hunyuan3d2025hunyuan3d,xiang2025structured}, our model demonstrates considerable potential for scene-level generation. 


The contributions of this work are as follows:
\begin{itemize}
    \item We observe that commonsense priors learned from videos can facilitate 3D generation, and we provide an initial exploration and validation of this video-driven 3D generative solution.
    \item We release \ourdataset{}, a large-scale 3D dataset paired with circumferential descriptions, containing 4 million 3D models, dense multi-view images, and detailed orbital text captions.
    \item We introduce \ourproj{}, a 3D multi-view generation model. It inputs both image and dense text and generates multi-view images, which supports downstream 3D Gaussian splatting and textured meshes reconstructions.
    \item We have open-sourced the dataset, code, the model weights, and the complete technical solution. We hope this initiative fosters algorithmic innovation in the public domain. 
\end{itemize}

\section{Related Work}

\subsection{3D Datasets}

The 3D dataset aims to construct a collection of 3D content encompassing various data formats to support the research and evaluation of intelligent algorithms such as reconstruction and understanding.
Existing 3D datasets are primarily constructed based on meshes or multi-view images.

Mesh-based 3D datasets, including Objaverse\cite{deitke2023objaverse}, Objaverse-XL\cite{deitke2023objaverse2025}, ShapeNet\cite{chang2015shapenet}, Thingi10K\cite{zhou2016thingi10k}, 3D-FUTURE\cite{fu20213d}, and OmniObject3D\cite{wu2023omniobject3d}, provide mesh models defined by vertices, edges, and faces. 
This format makes them highly suitable for tasks such as 3D rendering and geometric analysis. 
For example, the widely used GSO\cite{downs2022google} dataset falls into this category, which constructs 3D mesh structures based on scans of everyday objects. 
Additionally, the currently largest dataset, Objaverse-XL\cite{deitke2023objaverse2025}, also belongs to this category, containing over 10.2 million 3D objects sourced from a variety of origins, ranging from artificially designed models to photogrammetric scans, significantly exceeding its base version which composes of 800 thousand models\cite{deitke2023objaverse}. 
However, despite its large scale, the quality of its samples varies considerably. 
Many samples contain only limited geometric patches and lack sufficient visual textures and text annotation information. 
This limitation restricts the application of such datasets in highly controllable 3D generation tasks requiring precise perspective-aware descriptions.

On contrast, multi-view image-based 3D datasets, such as MVImgNet\cite{yu2023mvimgnet}, MVImgNet2.0\cite{han2024mvimgnet2}, and Co3D\cite{reizenstein2021common-CO3D}, provide 2D images or video frames capturing objects or scenes from multiple angles, supporting tasks such as Structure-from-Motion and 3D reconstruction. 
Among them, MVImgNet\cite{yu2023mvimgnet} includes 6.5 million frames from 219,188 videos across 238 categories, along with annotations such as masks and camera parameters. 
The further expanded MVImgNet2.0\cite{han2024mvimgnet2} covers approximately 300K objects with complete 360-degree views. 
Although these datasets provide rich multi-view data, their scale still lags significantly behind that of Objaverse-XL. 
Moreover, they also lack detailed text captions describing the appearance from specific perspectives.
For example, although Cap3D\cite{luo2023scalable} provides multi-view images, it provides short captions that describe the object’s overall appearance, thereby overlooking the variations that should inherently exist in descriptions across different viewpoints.
This hinders tasks requiring the integration of visual-text understanding, such as text-guided 3D or video generation.

In light of this, we propose the large-scale 3D dataset \ourdataset{}, designed to support the training of image-and-text-based 3D generation tasks. 
This dataset includes high-quality mesh models strictly selected from Objaverse-XL, as well as densely distributed multi-view rendered images surrounding the objects. 
More importantly, \ourdataset{} provides detailed text annotations at the surrounding multi-view level with an average length of 260 words, offering critical multi-modal supervision annotations for 3D generation tasks.

\subsection{3D Generation Models}

In recent years, the task of generating 3D content from text or a single image has advanced rapidly, underscoring the remarkable potential of AIGC to expand our repository of 3D digital assets. 
Existing research falls broadly into two paradigms.

The first paradigm, termed native 3D generation methods \cite{nichol2022point,jun2023shap,zhao2025hunyuan3d20}, can directly generate editable 3D assets represented by meshes or voxels from image or text prompts. 
These methods typically require training on large-scale datasets and employ feedforward inference mechanisms to achieve efficient generation. 
Although highly efficient and multi-view consistent, their performance heavily depends on the scale and quality of the training data, which often face accessibility and acquisition limitations.

Harnessing the remarkable generative power of large-scale pre-trained models in images \cite{rombach2022high,li2024image}, and videos \cite{openai-sora,zhang2025dropletvideo}, the second paradigm unifies 2D generative models to distill or reconstruct multi-view observations into 3D representations such as NeRF and 3DGS, yielding an efficient 3D generation framework.
In the line of combining image generation models, pioneering work DreamFusion \cite{poole2022dreamfusion} and subsequent studies \cite{tang2023make,yi2024gaussiandreamer,lin2023magic3d,chung2023luciddreamer,yi2024gaussiandreamerpro,wang2023prolificdreamer} employ techniques like Score Distillation Sampling (SDS) \cite{poole2022dreamfusion} or its improved variants to distill 3D information from 2D diffusion priors. 
To enhance multi-view consistency, more recent studies \cite{shi2023mvdream,liu2023zero,xu2023dmv3d} introduce conditional controls, such as camera embeddings \cite{shi2023zero123++,liu2023zero} and epipolar constraints \cite{huang2024epidiff}, to fine-tune pre-trained diffusion models. 
Meanwhile, other works \cite{hong2023lrm,tang2024lgm,xu2024instantmesh} propose to learn 3D via a two-stage pipeline: first generating intermediate multi-view images and then utilizing these images to assist in feedforward 3D reconstruction. However, the generated intermediate images in such methods are typically sparse, and even with the limited range, leading to the spatial consistency required for high-quality 3D reconstruction remains elusive.

More recently, video generation models \cite{blattmann2023stable,hong2022cogvideo} have effectively maintained inter-frame consistency by integrating temporal continuity, making them particularly suitable for generating coherent multi-view outputs. 
Inspired by this, several studies (e.g., V3D\cite{chen2024v3d}, SV3D\cite{voleti2024sv3d}, VideoMV\cite{zuo2024videomv}, etc. \cite{chen2024v3d,voleti2024sv3d,melas20243d,zuo2024videomv,yang2024hi3d,han2024vfusion3d}) suggest adopting fine-tuned video diffusion models to generate multi-view outputs for subsequent 3D generation. 
Although impressive generation results, these approaches fail to adequately leverage text supervision and lack explicit mechanisms to preserve the capabilities transferred from the video backbone.
Additionally, some methods \cite{huang2025edit360,li2024controllable,xu2024flexgen,melas20243d,lin2025kiss3dgen,yang2025tv} consider jointly modeling both text and image prompts to produce 3D assets with better appearance and geometry control. For example, IM-3D \cite{melas20243d} generates multi-view images from text and image pairs with the aid of a video diffusion model, followed by 3D reconstruction to generate high-quality 3D assets. 
Nevertheless, videos and video generation models are still underexplored in 3D generation due to simple text or image prompts. 
Therefore, in this work, we introduce a novel 3D generative model, \ourproj{}, which supports both image and dense text input, to further explore the potential of video modality in advancing 3D generation. Our model allows generate content that meets user needs with greater granularity,  and supports downstream 3D Gaussian splatting and textured meshes reconstructions.

\section{The \ourdataset{} Dataset}

\begin{figure}[htbp]
\centering
\includegraphics[width=1.0\linewidth]{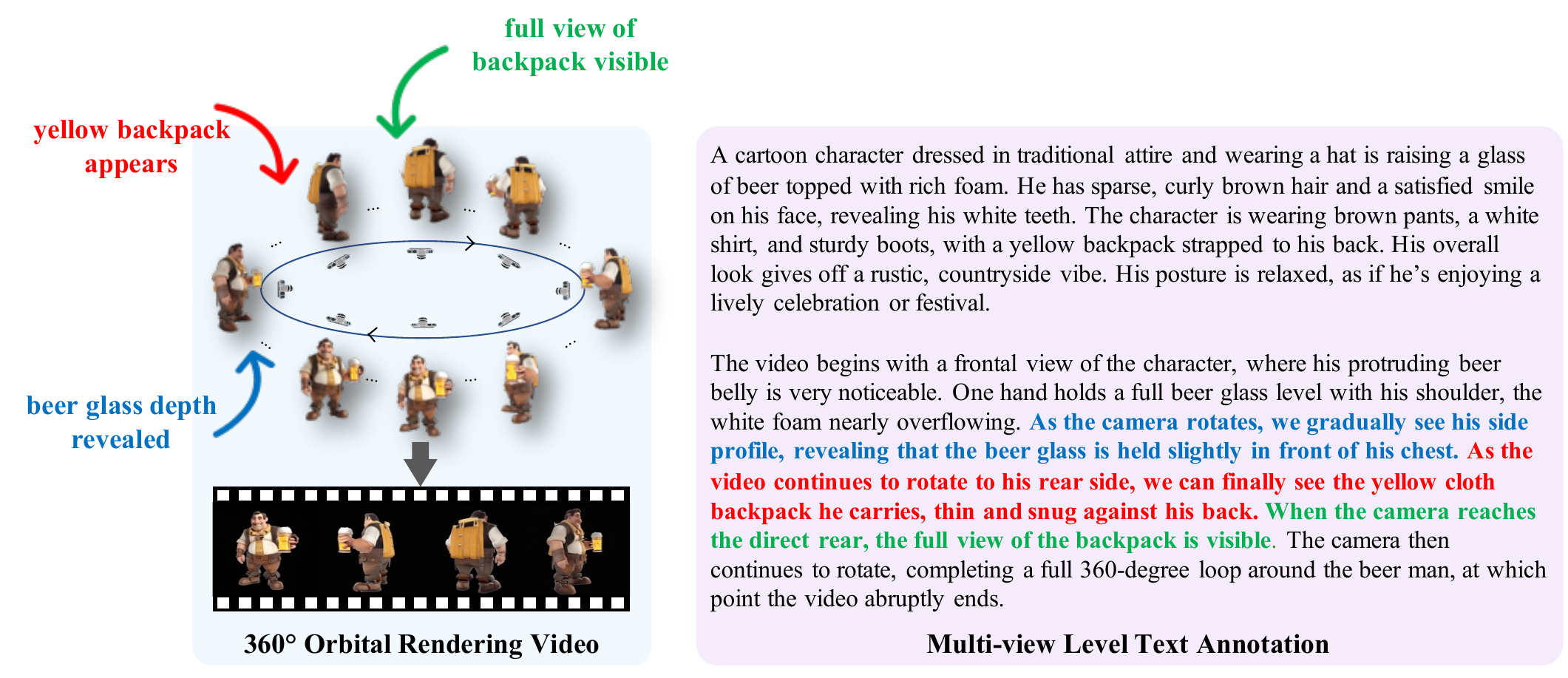}
\caption{\textbf{A sample from \ourdataset{} comprises a 85-frame multi-view rendered video and a fine-grained, multi-view-level text annotation.} The \textbf{\textcolor{blue!80!black}{blue}}, \textbf{\textcolor{red}{red}}, and \textbf{\textcolor{green!50!black}{green}} text illustrate viewpoint and appearance changes at three consecutive time steps, with corresponding left-side color markers indicating the specific moments of transformation.}
\label{fig:DatasetDemo}
\end{figure}

To take advantage of the commonsense priors acquired by video generation models trained in large-scale video data and thereby enhance the generalization capability of 3D generative models, we construct a novel dataset \ourdataset{} to bridge the gap between the video and 3D domains.
Specifically, a qualifying 3D sample must satisfy two criteria.
First, it should support dense, coherent multi-view sequences.
This ensures compatibility with the output interface of video generation models, facilitating the generation of spatially consistent 3D content.
Second, it should be accompanied by dense text annotations.
This provides stronger supervision, enabling the 3D generator to maximally retain the video model's semantic understanding of text. 
As the 3D generator inherits broader semantic knowledge from video data, this dense text supervision allows the generative model to effectively exploit this inherited capability.

\begin{table}[h]
\centering
\caption{\textbf{Comparison of \ourdataset{} and other 3D-language datasets.} \ourproj{} is a large-scale dataset of 4 million 3D objects, distinguished by its provision of dense 85-view renderings and precisely aligned, multi-view level dense text annotations for each object, which is novel among existing 3D datasets.
Note that \textbf{MVI} stands for Multi-view Images.}
\label{tab:Datasetcompare}
\small
\begin{tabular}{lccccccc}
\toprule
  & Year & Objects & Format & Images & Views & Annotations & Anno-Grain \\
\midrule
GSO\cite{downs2022google} & 2024 & 1K & Mesh & $-$ & $-$ & $-$ & $-$ \\
OmniObject3D\cite{wu2023omniobject3d} & 2024 & 6K & Mesh & $-$ & $-$ & $-$ & $-$ \\
Thingi10K\cite{zhou2016thingi10k} & 2016 & 10K & Mesh & $-$ & $-$ & $-$ & $-$ \\
3D-FUTURE\cite{fu20213d} & 2021 & 10K & Mesh & $-$ & $-$ & $-$ & $-$ \\
Objaverse-OA\cite{lu2025orientation} & 2025 & 14.8K & Mesh & $-$ & $-$ & $-$ & $-$ \\
ShapeNet\cite{chang2015shapenet} & 2015 & 51K & Mesh & $-$ & $-$ & $-$ & $-$ \\
Anymate\cite{deng2025anymate} & 2025 & 230K & Mesh & $-$ & $-$ & $-$ & $-$ \\
Objaverse\cite{deitke2023objaverse} & 2022 & 800K & Mesh & $-$ & $-$ & $-$ & $-$ \\
TexVerse\cite{zhang2025texverse} & 2025 & 858K & Mesh & $-$ & $-$ & 3 sentence & Object Level \\
Objaverse-XL\cite{deitke2023objaverse2025} & 2023 & \textbf{10.2M} & Mesh & $-$ & $-$ & $-$ & $-$ \\
\midrule
CO3D\cite{reizenstein2021common-CO3D} & 2021 & 19K & MVI & 1.5M & 79 & $-$ & $-$ \\
MV-Video\cite{jiang2024animate3d-MV-Video} & 2024 & 53K & MVI & 88M & 48 & $-$ & $-$ \\
uCO3D\cite{liu2025uncommon} & 2025 & 170K & MVI & 34M & 200 & $-$ & Object Level \\
G-Objaverse\cite{zuo2024high} & 2024 & 200K & MVI & 30M & 38 & $-$ & $-$ \\
MVImgNet\cite{yu2023mvimgnet} & 2023 & 220K & MVI & 6.5M & 30 & $-$ & $-$ \\
MVImgNet2.0\cite{han2024mvimgnet2} & 2024 & 300K & MVI & 9M & 30 & $-$ & $-$ \\
Objaverse++\cite{lin2025objaversepp} & 2025 & 500K & MVI & 20M & 40 & $-$ & $-$ \\
Cap3D\cite{luo2023scalable} & 2023 & 791K & MVI & 6M & 8 & $<$15.0 words & Object Level \\
\textbf{Droplet3D-4M} & 2025 & \textbf{4M} & MVI & \textbf{900M} & \textbf{85} & \textbf{260.0} words & Multi-view Level \\
\bottomrule
\end{tabular}
\end{table}

Accordingly, our proposed dataset, \ourdataset{}, comprises dense multi-view rendered videos and fine-grained, multi-view-level text annotations, as illustrated in Fig. \ref{fig:DatasetDemo}. 
Each rendered video consists of an 85-frame image sequence captured from uniformly distributed 360° orbital viewpoints. 
The angular difference between adjacent frames is strictly controlled to be within 5°, ensuring video coherence, which is a critical factor In training video-driven generative models. 
In terms of text annotations, we provide dense descriptions with an average length of 260 words, far exceeding those in existing 3D datasets. 
More importantly, the annotations not only cover holistic appearances such as shape and style but also specifically describe appearance variations induced by changes in viewpoint. 
For example in Fig. \ref{fig:DatasetDemo}, the second paragraph of the text annotation details the side and rear features of the figurine, describing that the yellow backpack becomes partially visible from a side view and is only fully revealed from the back.
This fine-grained, view-aware annotation paradigm provides unprecedented supervisory signals for the 3D generative model, effectively guiding and preserving the backbone network’s capacity for complex semantic understanding. 
In contrast to current prevailing 3D datasets, which typically lack text annotations or provide only brief descriptions at object-level, \ourdataset{} achieves a significant enhancement in information richness, which has been illustrated in Tab. \ref{tab:Datasetcompare}.

\begin{figure}[htbp]
\centering
\includegraphics[width=1.0\linewidth]{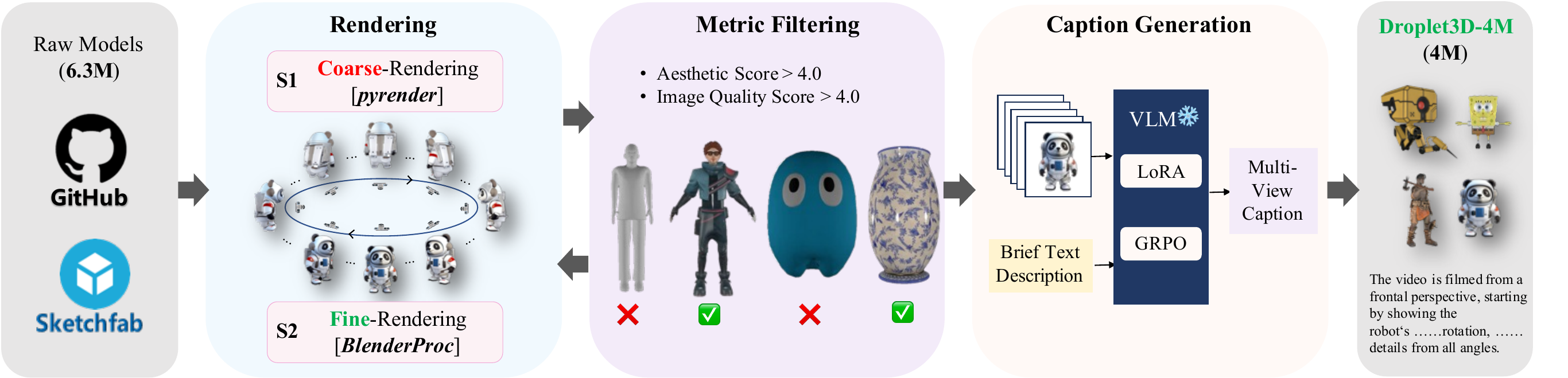}
\caption{\textbf{The pipeline we proposed to curate the \ourdataset{} dataset.} The proposed pipeline encompasses a three key stages: multi-granularity rendering, filtering, and caption generating.}
\label{fig:DatasetPipeline}
\end{figure}

For the construction of \ourdataset{}, we propose a data pre-processing pipeline that balances quality and efficiency, and construct a new dataset, \ourdataset{}.
This pipeline employs adaptive sampling techniques instead of resource-intensive rendering during the initial screening phase, followed by targeted high-fidelity rendering only for validated assets.
Compared to conventional workflows, this approach reduces computational overhead by a factor of $4$ to $7\times$ while generating metadata-rich outputs of superior quality for multimodal learning tasks.
The pipeline consists of three key parts: multi-view video rendering, image evaluation metric filtering, and multi-view-level caption generation, as shown in Fig. \ref{fig:DatasetPipeline}.
Note that the raw 3D models we collected are from Objaverse-XL\cite{deitke2023objaverse2025}, comprising 6.3 million models sourced from GitHub and Sketchfab.

\subsection{Multi-View video Rendering}

We designed both coarse and fine rendering strategies for multi-view videos to mitigate the explosive computational consumption caused by rendering all data directly using the Blender engine.
For the coarse rendering, we initially rendered the 6.3 million 3D models sourced from Objaverse-XL at low quality using the pyrender library.
For the fine rendering, we employed BlenderProc to render the filtered 4 million models that met the quality and aesthetic score criteria.

The fundamental rendering hyperparameters were kept consistent across both rendering processes. 
The perspective camera configuration was set to a 50-degree vertical field of view and maintained a fixed aspect ratio of 1.0.
Prior to rendering, the 3D meshes underwent geometric normalization, including aligning the centroid to the world origin, uniformly scaling to a unit maximum extent, and optional axial rotation.
The scene lighting utilized a multi-light setup: a directional ambient light with an intensity of 2.0 provided base illumination, supplemented by three positional point lights, each with an intensity of 8.0, strategically placed on the left, right, and rear of the object at a distance of 2 units.

The coarse rendering process leveraged PyRender's off-screen rendering capabilities and EGL backend acceleration, while the fine rendering process utilized Blender's EEVEE renderer to enhance quality.
For the coarse rendering process, only 8 views were rendered for metric calculation and filtering.
For the high-quality rendering, 85 camera positions were uniformly distributed along a circular trajectory with a fixed 0° elevation angle, with radial distances randomly varying between 1.6 and 2.0 units. All cameras were focused on the scene origin, with the $+Y$ axis serving as the upward reference direction.

\subsection{Image-Assessment Metric Filtering}

We filtered the dataset by selecting high-quality multi-view rendering videos based on aesthetics and image quality.
We utilize the publicly available LAION aesthetics model\cite{schuhmann2022laion} to compute aesthetic scores and the DOVER-Technical model\cite{wu2023exploring-dover} to evaluate image quality.
Only clips surpassing predefined thresholds are retained.
Notably, nearly 77\% of clips achieve an aesthetic score above 4.0, while approximately 81\% exceed a score of 4.0 in image quality, underscoring the dataset's high visual fidelity.

\subsection{Multi-View consistent captions Generation}

\subsubsection{Supervised-finetuning VLM captioner}  

The proposed framework employs an AI-driven approach to convert video content into detailed textual descriptions, significantly reducing the need for human annotation. 
Traditional video captioning methods often produce brief outputs that fail to capture nuanced visual elements across different perspectives. 
To address this, our methodology incorporates a dual-phase enhancement mechanism. 
First, we construct a curated video dataset featuring comprehensive annotations that precisely track object states, camera movements, and – most importantly – variations in detail levels across viewpoint transitions. 
These preliminary descriptions are then linguistically refined using GPT-4 to ensure grammatical precision and stylistic consistency, resulting in high-quality training material.

For the technical implementation, we fine-tuned multiple cutting-edge multimodal models within the 7-9 billion parameter spectrum, considering both performance and efficiency.
The generated annotations exhibit variations in detail levels across viewpoint transitions – including precise camera operations, object-surface changes, illumination changes, and environmental characteristics. 
Such meticulous documentation offers vital training cues for keeping the multi-view correspondence, especially concerning appearance changes. 
Quality assurance measurements indicate that the automated captions match or surpass human-generated annotations in terms of multi-view correpondence.

\subsubsection{Enhancing MLLMs for Multi-View Captioning via GRPO}

\begin{figure}[ht]
    \centering
    \includegraphics[width=1.0\linewidth]{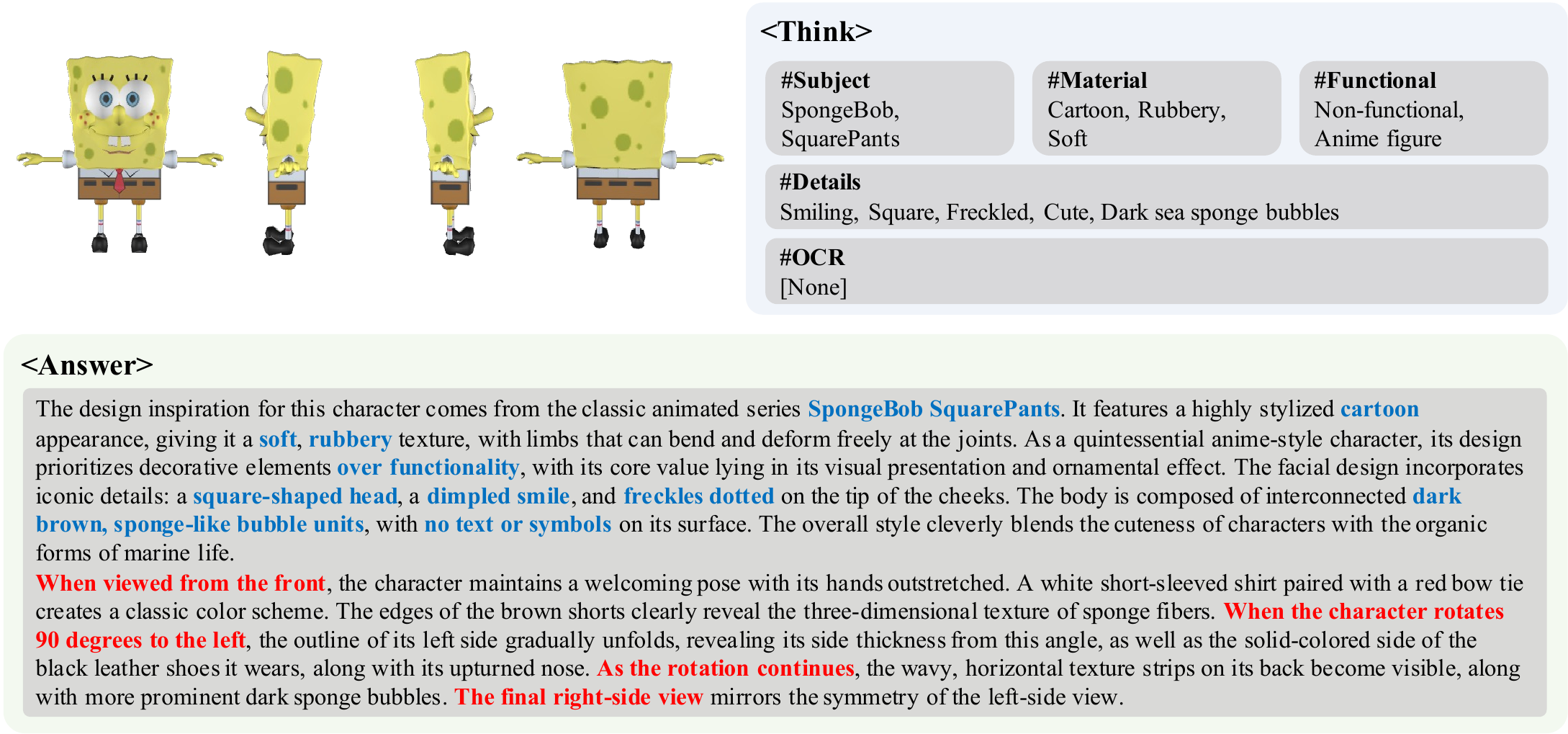}
    \caption{\textbf{An example of reinforcement learning samples for GRPO.} We constructed a constrained model to annotate the \textbf{think} and \textbf{answer} processes. For the think process, we provided annotations on themes, materials, functions, details, OCR, etc. For the answer process, we identified two types of scoring points based on dense multi-view level captions: scoring points marked in \textcolor{blue}{\textbf{blue}} denote those aligned with the Think process, while those marked in \textcolor{red}{\textbf{red}} indicate detailed descriptions of perspective changes.}
    \label{fig:Grpo}
\end{figure}

Recent studies have demonstrated that test-time scaling can significantly enhance the reasoning abilities of large-scale language models. 
This technique has been successfully implemented in state-of-the-art systems such as OpenAI's o1\cite{jaech2024openai} and Deepseek-R1\cite{guo2025deepseek}, both of which achieve strong performance on challenging tasks involving mathematical reasoning and programming. 
However, vision-language models (VLMs) fine-tuned using conventional approaches frequently exhibit problematic behaviors, including incorrect object type predictions, misleading functional descriptions, and hallucinations of non-existent visual details. 
To mitigate these issues, the GRPO algorithm\cite{shao2024deepseekmath} has proven particularly effective when applied to supervised fine-tuning for specialized vision tasks. 
Empirical results confirm its advantages across multiple domains, including visual question answering\cite{zhao2025mmvu}, spatial relationship understanding\cite{shen2025vlm}, and temporal sequence analysis in video content\cite{feng2025video}.

Therefore, we designed the VLM model enhancements throught the second stage GRPO Reinforcement learning. 
We designed five dimensions to ensure the output descriptions meet our requirements for multi-perspective image captioning. 
They are Subject(Judging the type of the object), material(what the object is made of), Functional(what the object is used for), Details(changes due to the camera motion), OCR(whether the visual text during the view change). 
Also, similar to Deepseek-R1, we designed \texttt{``<think>dimensions</think><answer>caption</answer>''} as the output format. 
The dimensions contain the descriptions of the object for the five dimensions above. 
The caption part is the fluent mutli-view caption for this sample, combining the dimension contents above with the view changes. 
Fig. \ref{fig:Grpo} shows an example. 

\begin{figure}[htbp]
\centering
\includegraphics[width=0.9\linewidth]{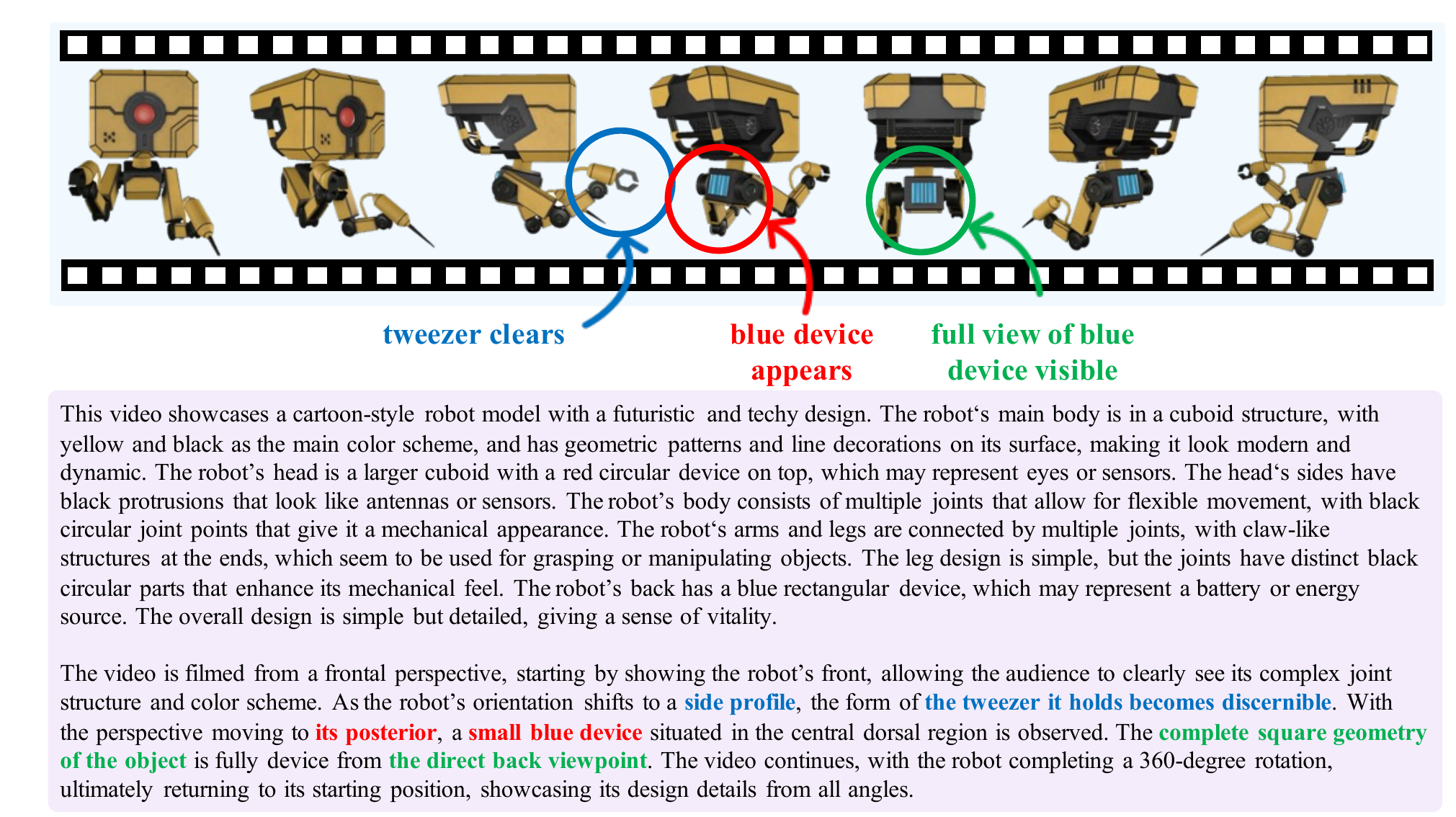}
\caption{\textbf{An caption illustration from \ourdataset{}.} Each 3D model is ultimately rendered into an encompassing multi-view video comprising 85 frames with uniformly distributed angles. The caption typically consists of two paragraphs: the first describes the overall style and key details of the object, while the second provides a multi-view-level description, emphasizing the appearance variations induced by the circular capture. Expressions related to viewpoint changes are highlighted in different colors.}
\label{fig:DatasetSample}
\end{figure}

For the Reward funtions, we mainly focus on the content metric and the multiview-match format metric. 
The content metric formula is as:

\begin{equation}
    P = \frac{1}{N}\sum_{i=1}^{N} \text{sign}(i) \cdot \text{sim}(p_{i}, g_{i})
\end{equation}

\begin{equation}
    R = \frac{1}{M}\sum_{i=1}^{M} \text{sign}(i) \cdot \text{sim}(p_{i}, g_{i})
\end{equation}

\begin{equation}
    F1 = \frac{2 \cdot P \cdot R}{P + R}
\end{equation}

The multiview-match format metric is as:
\begin{equation}
M = \min\left(1.0,\ \frac{\sum \mathrm{matches}(\mathrm{pattern}:MV_\mathrm{pattern})}{5}\right)
\end{equation}

The Multi-view pattern can be described as the more key words, like front, circle-around, side, back, self-rotation, 360-degree, appear, the more pattern award will be obtained.
The final reward is determined as: $Reward = F1 + M$.
Ultimately, captions that are insufficiently detailed are compelled to incorporate additional information. 
As illustrated in Fig. \ref{fig:DatasetSample} with samples from \ourdataset{}, it is evident that the captions contain substantial details regarding the appearance and perspectives.

\subsection{Data Overall and Statistics}

In conclusion, the constructed \ourdataset{} dataset comprises a meticulously curated collection of 4 million fully-covered multi-view video sequences, encompassing approximately 900 million individual frames with a total runtime exceeding 8,000 hours.
Its unprocessed raw 3D models were collected from Objaverse-XL\cite{deitke2023objaverse2025}, comprising 6.3 million models sourced from GitHub and Sketchfab.
This comprehensive visual dataset has been publicly released with the aim of advancing research in spherical vision systems and multi-view analysis.
It should be noted that due to the origin of some source models being sampled from the internet, the dataset is published under the CC BY-NC-SA 4.0 license.
Researchers are advised that although the material is freely available for academic research, all content within it remains subject to the original distribution permissions, restricting its use to non-commercial purposes only.

\begin{figure}[htbp]
\centering
\includegraphics[width=0.9\linewidth]{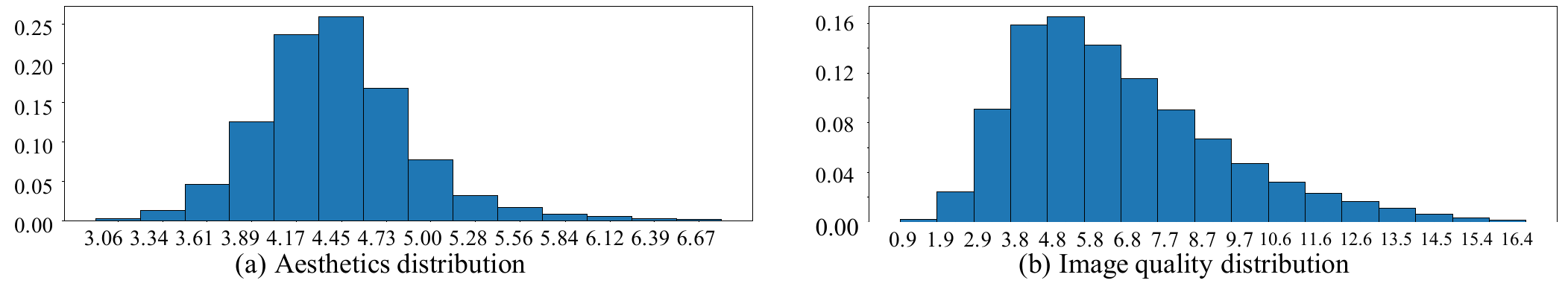}
\caption{\textbf{The aesthetics distribution and the image quality distribution of \ourdataset{}.} These distributions demonstrate that our dataset achieves \textbf{high scores} in both aesthetics and image quality, indicating an overall \textbf{high-quality standard} for the dataset.}
\label{fig:DatasetStatics}
\end{figure}

In addition, we computed the aesthetic and quality scores of the rendered videos in the dataset, as shown in Fig. \ref{fig:DatasetStatics}.
As previously mentioned, \ourdataset{} ultimately retained only those samples with both scores above 4.0.


\section{The Droplet3D Model}

\begin{figure}[h]
\centering
\includegraphics[width=0.88\linewidth]{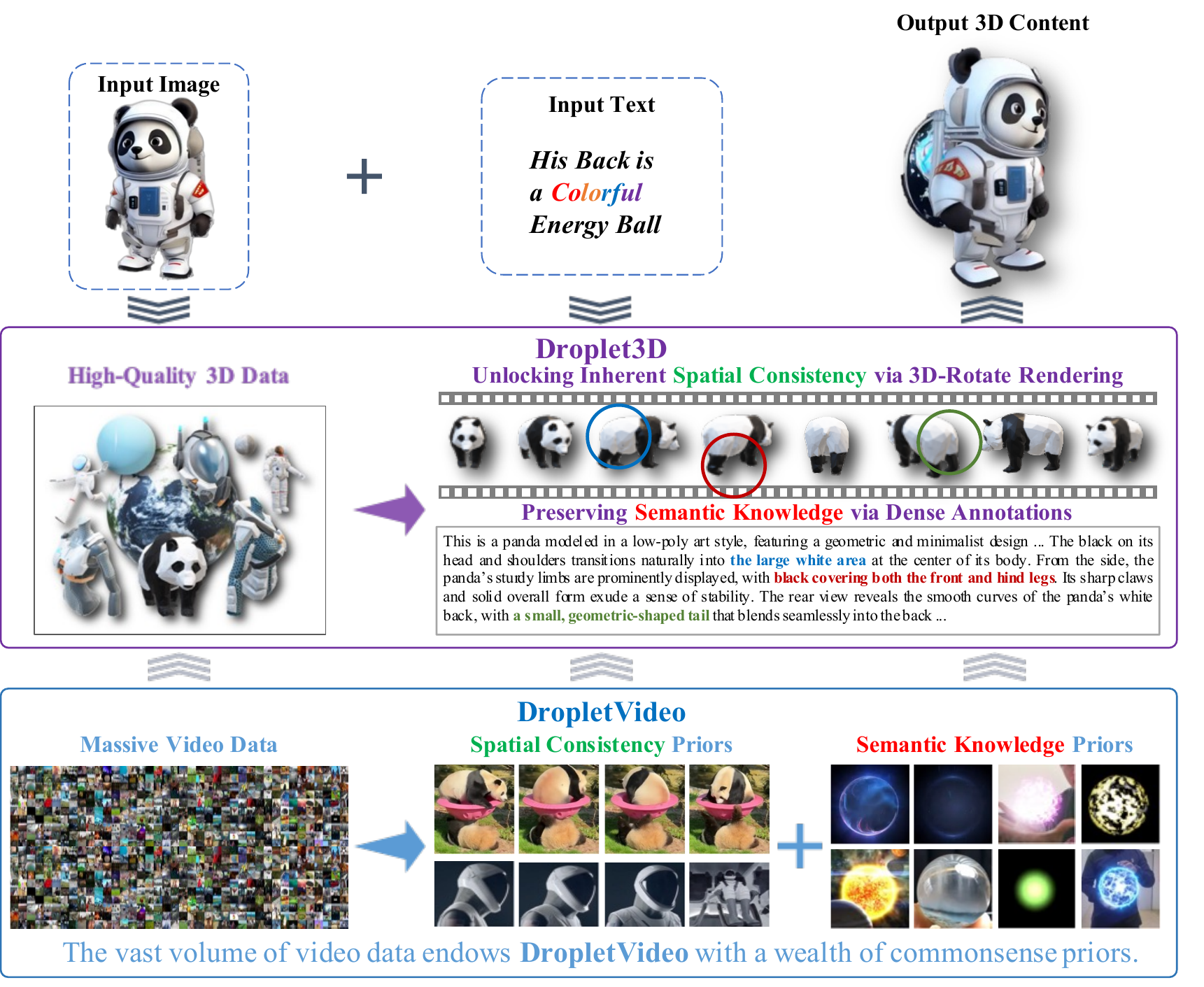}
\caption{\textbf{\ourproj{} Framework: Inheriting spatial-semantic priors from massive videos for high-fidelity 3D generation.} \ourproj{} employs DropletVideo as its video backbone, effectively leveraging its commonsense priors of spatial consistency and semantic knowledge. This enables a novel 3D object generation paradigm conditioned on the joint input of an initial image and dense text, while achieving superior generalization performance.}
\label{fig:Droplet3DThought}
\end{figure}

To inherit the spatial consistency and semantic knowledge priors that are readily accessible to video models into 3D generation algorithms, we designed and trained \ourproj{}. 
To this end, we specifically employ multi-view rendered videos from \ourdataset{} to interface with the video backbone model. 
Furthermore, we use multi-view level annotations to provide fine-grained supervision for model fine-tuning, thereby preserving the model’s cross-modal semantic understanding capability. 
The architecture of Droplet3D is illustrated in Fig. \ref{fig:Droplet3DThought}.

\begin{figure}[h]
\centering
\includegraphics[width=1.0\linewidth]{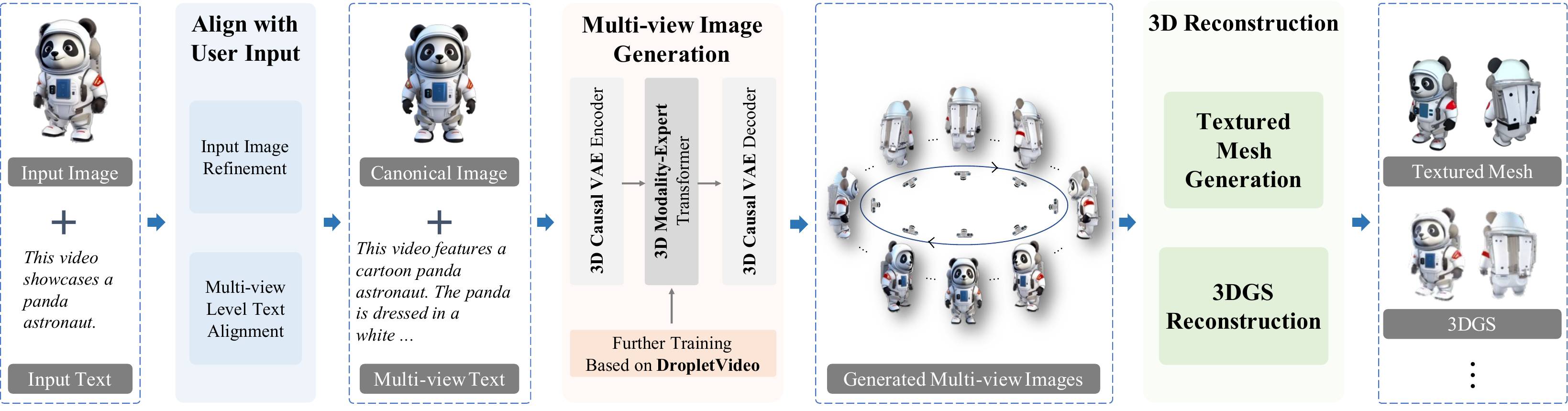}
\caption{\textbf{Overview of the \ourproj{} Techniques.} To enhance performance, \ourproj{} supports a alignment module to transfer the user input to fit the model. Subsequently, the aligned text and image features are fed into the backbone network to generate multi-view images with 3D consistency. At the end, the created multi-view content can be integrated into reconstruction modules for various 3D modalities, thereby generating production-ready 3D assets.}
\label{fig:Droplet3DArch}
\end{figure}

The technical details of the 3D generation process during \ourproj{} is illustrated in Fig. \ref{fig:Droplet3DArch}.
For any given text or image prompt, \ourproj{} first aligns it with the model. 
Initially, we expand the user's input requirements based on a lightweight large language model, rewriting them into dense textual descriptions that follow the same distribution as the multi-view level captions in the \ourdataset{} dataset. 
Simultaneously, to accommodate inputs from arbitrary viewpoints, we designed a anonical viewpoint alignment module to adjust the perspective of the input images. 
For the backbone network, we introduce a 3D causal VAE to achieve implicit space encoding and decoding of the video, and employ a multi-modal diffusion transformer to facilitate the fusion of text and video modality features while constraining their independence.

\ourproj{} demonstrates creativity in the process of generating surrounding multi-view videos based on images and text prompts. 
As an application, we adapted and integrated various converters to reconstruct different 3D modalities, such as textured mesh and Gaussian splatting, serving as applications of the content generated by \ourproj{}.

\subsection{Multi-View Image Generation}

The generation of multi-view images, or surround-view videos, is a core step in \ourproj{}, and it is crucial for creative 3D generation based on images and text.
From our perspective, this step can be viewed as a bridge that transfers the general capabilities of video models to 3D generation.

\begin{figure}[h]
\centering
\includegraphics[width=0.8\linewidth]{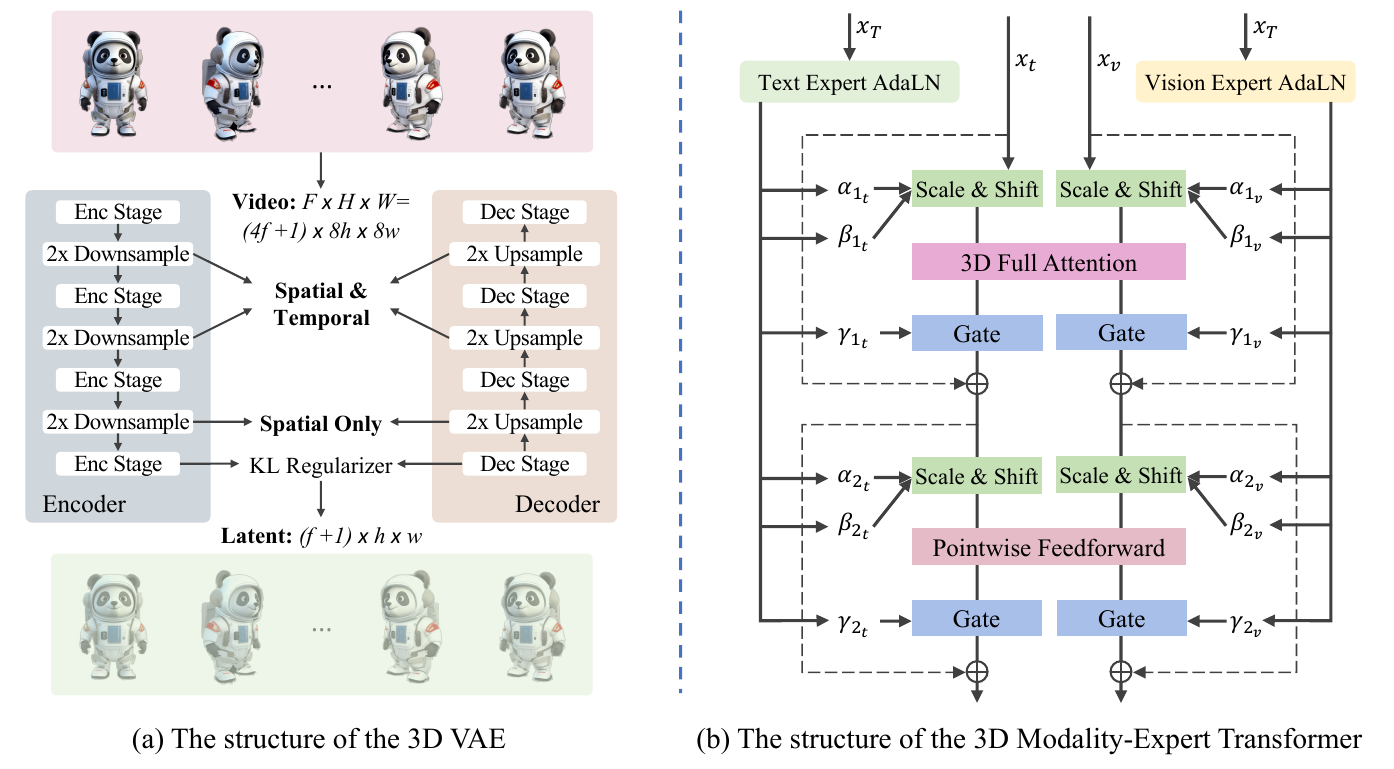}
\caption{\textbf{The surrounding multi-view video generation backbone of \ourproj{}}. It consists of two components: a 3D causal VAE and a vision-text modality-expert Transformer architecture.}
\label{fig:Backbone}
\end{figure}

\noindent{\textbf{3D Causal VAE}}

Within the backbone network of \ourproj{}, videos are initially processed by a 3D causal Variational Autoencoder (VAE) to be transformed into latent space features.
We extend the 3D Causal VAE using a VAE framework coupled with a 3D structural architecture, designed for encoding and decoding video frames.
Unlike traditional autoencoders, the outputs of the VAE's encoder and decoder are modeled as parameterized probability density distributions.
Yang et al. \cite{yang2024cogvideox} applied 3D convolutions to video reconstruction, demonstrating that a 3D structure can effectively mitigate the flickering artifacts often present in reconstructed videos.
Inspired by this, we employ a 3D causal VAE to enhance the computational efficiency of \ourproj{}, while ensuring the generated videos possess temporal continuity and stability.
Simultaneously, on the long-text side, we extract linguistic features based on the encoder structure of the T5 model\cite{raffel2020exploring}, feeding these features concurrently into the subsequent 3D Modality-expert Transformers.

\noindent{\textbf{3D Modality-Expert Transformer}}

Within the Transformer backbone network, we employ a 3D multi-modal attention mechanism to simultaneously process both text and video inputs.
Specifically, for the video input, given that 3D full attention is a technique that has evolved alongside the widespread adoption of Transformers in computer vision, we incorporate 3D positional embeddings within the Transformer architecture.
Concurrently, the text input is encoded through a conventional Transformer encoder to facilitate the smooth fusion of multi-modal information.
Compared to previous decoupled approaches, this integrated methodology enables a more effective capture of dynamic changes within the video and enhances the generated content's performance in terms of semantic consistency and diversity.

\noindent{\textbf{Further Training Based on DropletVideo}}

Given that the fundamental rationale for introducing the video generation step is to incorporate the general capabilities of video generation models into 3D asset generation, we proceed with further training based on an appropriate video generation backbone model.
DropletVideo\cite{zhang2025dropletvideo} is a video generation model that considers integral spatio-temporal consistency, supporting the generation process given an initial frame image and dense text input. 
This aligns formally with the multi-view image generation pipeline of \ourproj{}. 
Additionally, because its training data includes large-scale video segments constrained by spatial consistency, such as certain street scene or character surround shots, it exhibits strong potential for 3D consistency.
Therefore, we conduct secondary training based on its weights to fit the latent space feature distributions for 3D surround-view generation.

\subsection{User Input Alignments}

\subsubsection{Multi-View-Level Text Alignment}

\begin{figure}[h]
\centering
\includegraphics[width=1.0\linewidth]{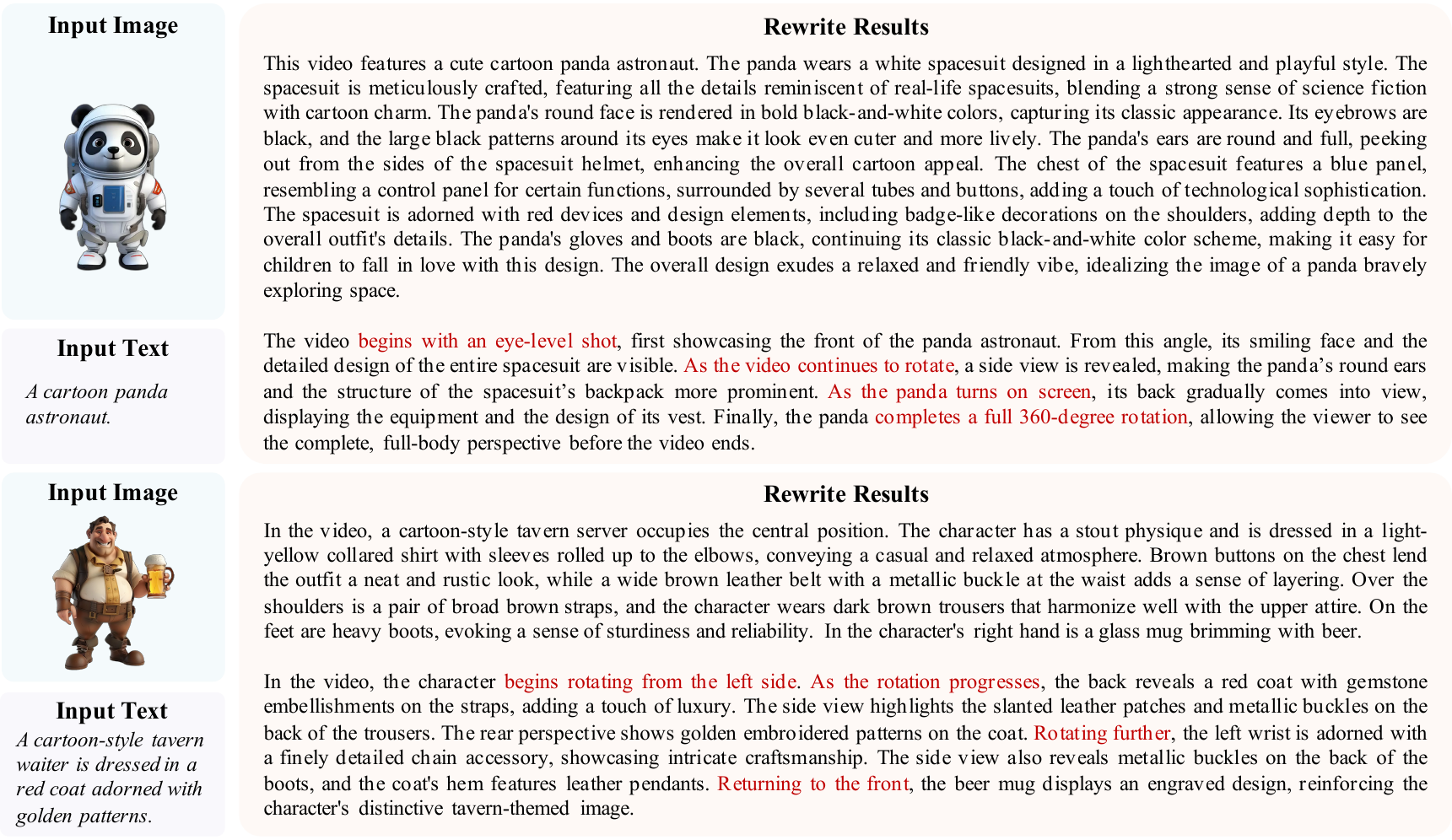}
\caption{\textbf{An illustration of rewritten samples.} Text describing viewpoint changes is highlighted in \textcolor{red}{red}. \textcolor{green!50!black}{Green} text indicates how user-provided brief or detailed descriptions have been rewritten into a form consistent with the training samples, showcasing their appearance across different viewpoints to provide more specific design requirements.}
\label{fig:RewriteSample}
\end{figure}

To effectively align the variations in linguistic style and length of user-provided prompts and to provide detailed guidance for the surround-view video generation of the target object, we implement a dense prompt generation preprocessing step.
Specifically, considering the superior performance of large language models in tasks such as text reasoning and image summarization, we employ the LoRA \cite{hu2021lora} technique to fine-tune an open-source model via instruction tuning. Experimental results indicate that approximately 500 in-domain samples are sufficient to achieve the desired fine-tuning level.
This module is designed to rewrite user prompts while preserving the original semantics. 
It transforms the prompts into a standardized information structure similar to the captions used during training. 
Similar to \ourdataset{}, the rewritten descriptions typically consist of two denser paragraphs. 
The first paragraph describes the overall style of the object and as many details as possible, while the second paragraph elaborates on multi-view level details based on the user's prompt, as illustrated in Fig. \ref{fig:RewriteSample}.
Users can review or modify the generated text for proofreading before inputting it into \ourproj{}.
Besides, this module supports multiple languages.

\subsubsection{Image Perspective Alignment}

\begin{figure}[h]
\centering
\includegraphics[width=1.0\linewidth]{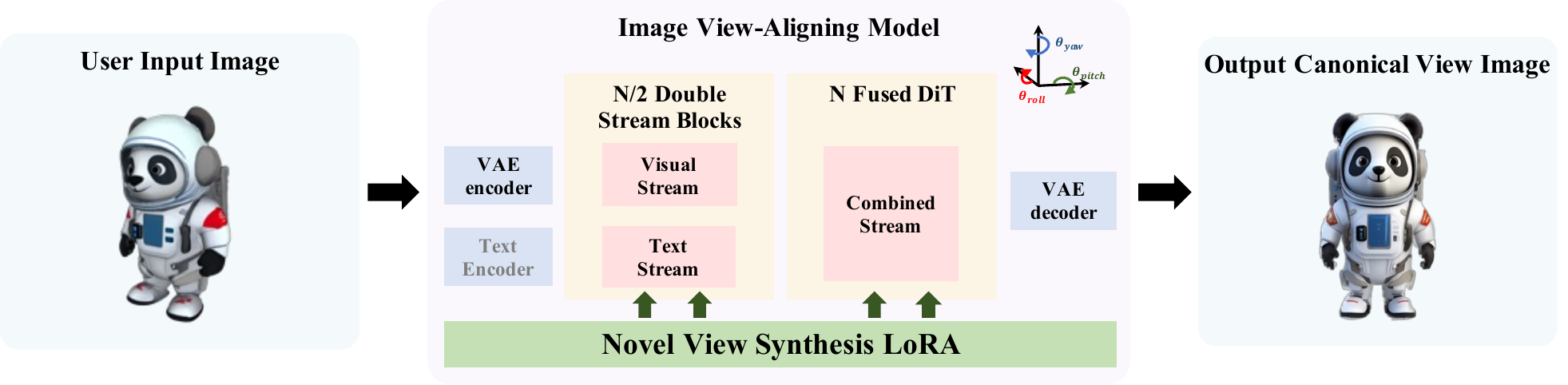}
\caption{\textbf{Image alignment module converts users' arbitrary inputs into canonical perspectives.} It employs LoRA fine-tuning based on the FLUX.1-Kontext-dev model to rectify arbitrary input viewpoints into canonical perspectives, such as front, back, left, and right.}
\label{fig:CVA}
\end{figure}

3D generation methods typically achieve optimal performance only when provided with a canonical view, demonstrating poor adaptability to arbitrary input perspectives. 
This imposes strict requirements on user input. 
For instance, when a user aims to perform 3D generation from an image captured at a rotated angle, the performance of existing generation algorithms significantly degrades. 
Motivated by this limitation, we introduce a view alignment module in \ourproj{}, enabling it to be seamlessly invoked with images from any viewpoint.
This module fine-tunes the input view to one of the canonical viewpoints: front, left, right, or back.

We selected some samples with high image-quality from the \ourdataset{} dataset and collected some AIGC generated samples from Internet, making a small dataset containing 200 examples. Then, we constructed the canonical viewpoints generation dataset from the small dataset. 
For each example, we manually selected four standard orthogonal views as \textbf{ground truth}, and constructed image pairs using initial images rendered by rotating $[15,\ 40]$ degrees clockwise or counterclockwise from the chosen ground truth viewpoint.
Finally,  we frame this task as an image editing problem, with the fixed editing prompt: \textit{``Convert this image to the closest canonical view, such as front, left, right or back.''}

For the model selection and fine-tuning, we selected the FLUX.1-Kontext-dev \cite{labs2025flux} as the base model, training it with
Low-Rank Adaptation (LoRA)\cite{hu2021lora}.
As shown in Fig. \ref{fig:CVA}, the model architecture employs a hybrid design combining dual-stream and single-stream blocks. In the dual-stream configuration, separate weight matrices process image and text tokens independently, with cross-modal fusion achieved through attention mechanisms applied to concatenated token sequences. 
Following dual-stream processing, the integrated token sequence undergoes joint encoding via successive single-stream blocks before final text token pruning and image token decoding.

\section{Implementation and Experiments}

\subsection{Implementation Details}

We employ the parameters of the pre-trained \texttt{DropletVideo-5B}\cite{zhang2025dropletvideo} model as the weight initialization for \ourproj{}.
Similar to DropletVideo, we employ \texttt{t5-v1\_1-xxl}\cite{raffel2020exploring} as the text encoder, and set the maximum token length to 400 instead of 226, to accommodate longer captions.
The model architecture is based on the MMDiT series\cite{esser2403scaling}, consisting of 42 layers, with 48 attention heads per layer, and each head has a dimension of 64.
The time step embedding dimension is set to 512.
For optimization, we use \texttt{Adam}\cite{kingma2014adam} with a weight decay of $3 \times 10^{-2}$ and an eps value of $1 \times 10^{-10}$.
The learning rate is set to $2 \times 10^{-5}$. The number of sampled frames ((N)) is fixed at 85.
The model is trained using the \texttt{bfloat16} mixed precision method, similar to the DeepSpeed framework\cite{rasley2020deepspeed}.

During inference, the classifier-free guidance scale is set to 6.5 to enhance the motion smoothness of the generated surround-view videos.
When trained on the \ourdataset{} dataset, our model supports an image resolution of 512.

For the Canonical View Alignment training, The LoRA rank (network dimension) was set to 128. The learning rate is 1e-4 and the optimizer is AdamW8bit\cite{dettmers20218}.

\subsection{Evaluation}

\subsubsection{TI-to-3D Quantitative Evaluation}

We conducted a quantitative comparative analysis based on the GSO dataset\cite{downs2022google}.
To validate the accuracy of the generated content by \ourproj{} given a single image and text prompt, we compared it with other TI-to-3D methods, including LGM\cite{tang2024lgm}, MVControl\cite{li2024controllable} as shown in Tab. \ref{tab:ExpTI23D}.

Given the bias in the coverage of the GSO dataset, we performed down-sampling and selected a subset of 200 samples for the evaluation.
The sampled instances can cover all categories in the original dataset while ensuring, to the greatest extent possible, that their distribution is confirmed uniform.

\begin{figure}[h]
\centering
\includegraphics[width=1.0\linewidth]{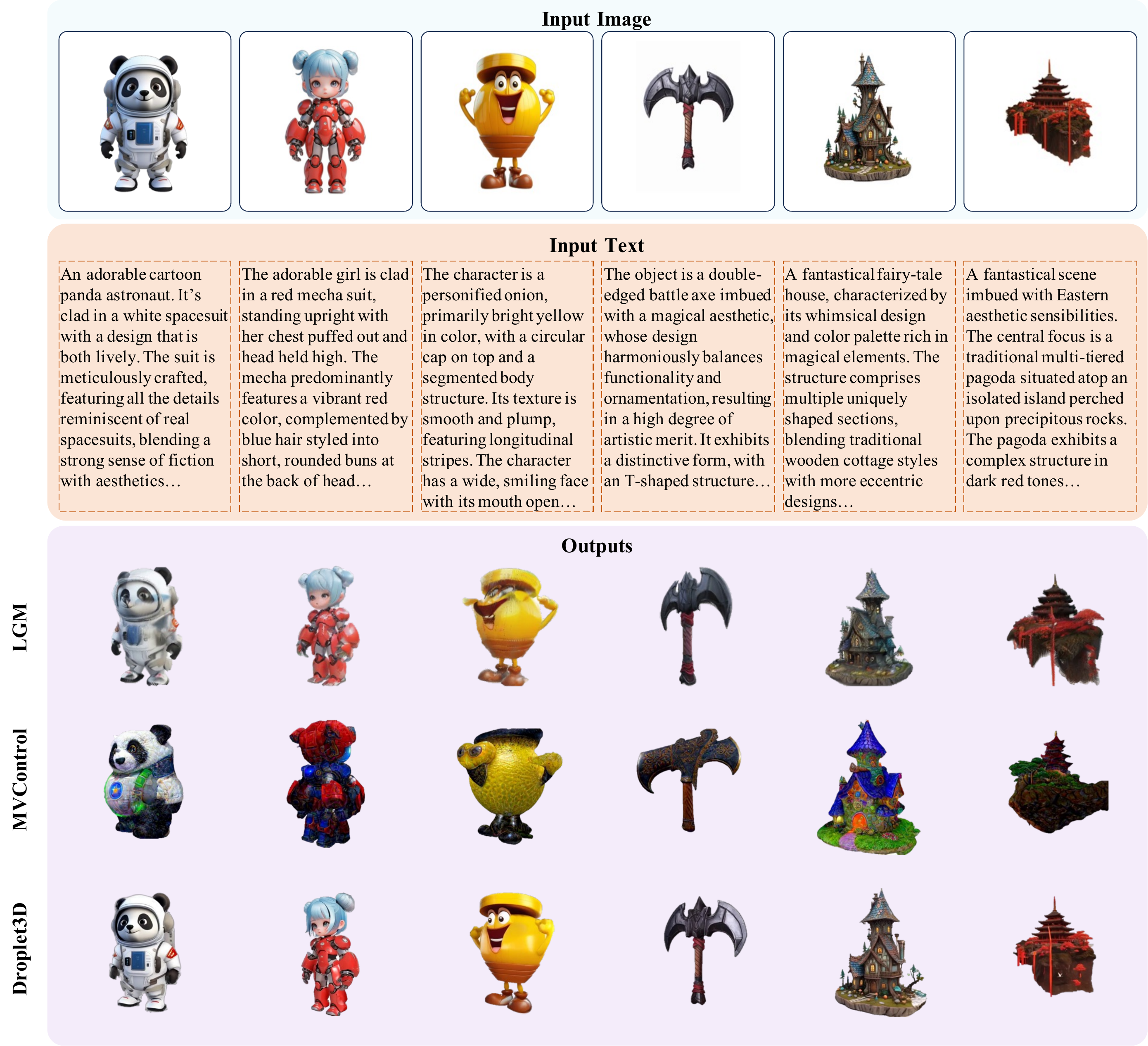}
\caption{\textbf{Comparison of generated content from different generative methods that support simultaneous image and text input.} The experimental cases demonstrate that the generation performance of \ourproj{} is significantly superior to that of the two baseline methods.}
\label{fig:CompareTI23D}
\end{figure}

\begin{table}[ht]
\centering
\caption{\textbf{Quantitative comparison of \ourproj{} with other 3D generation methods that simultaneously support both text and image inputs.} \ourproj{} outperforms the other two methods.}
\label{tab:ExpTI23D}
\small
\begin{tabular}{lcccccc}
\toprule
  & PSNR ($\uparrow$) & SSIM ($\uparrow$) & LPIPS ($\downarrow$) & MSE ($\downarrow$) & CLIP-S ($\uparrow$) \\
\midrule
LGM\cite{tang2024lgm} & 21.38 & 0.84 & 0.137 & 0.011 & 0.737\\
MVControl\cite{li2024controllable} & 22.31 & 0.88 & 0.31 & 0.009 & 0.61\\
\textbf{Droplet3D (Ours)} & 28.36 & 0.76 & 0.03 & 0.0017 & 0.866\\
\bottomrule
\end{tabular}
\end{table}

As shown in Tab. \ref{tab:ExpTI23D}, our model significantly outperforms the other two methods in terms of generation performance. 
On the one hand, metrics such as PSNR and SSIM confirm that the content generated by \ourproj{} is of higher quality and closer to the ground truth. 
On the other hand, it demonstrates stronger semantic understanding capabilities, as evidenced by the CLIP scores in the last column. 
We attribute this to our use of T5 as the text encoder and extensive pre-training on video data.

Fig. \ref{fig:CompareTI23D} presents a qualitative comparison between \ourproj{} and two other methods, LGM and MVControl, both of which also support simultaneous input of image and text. 
The results demonstrate that our method outperforms the others in terms of aesthetic quality, precision, and consistency between the generated content and the input image.
This further validates the superiority of our \ourproj{} in the TI-2-3D task.


\subsubsection{Ablation Study}

\begin{figure}[h]
\centering
\includegraphics[width=1.0\linewidth]{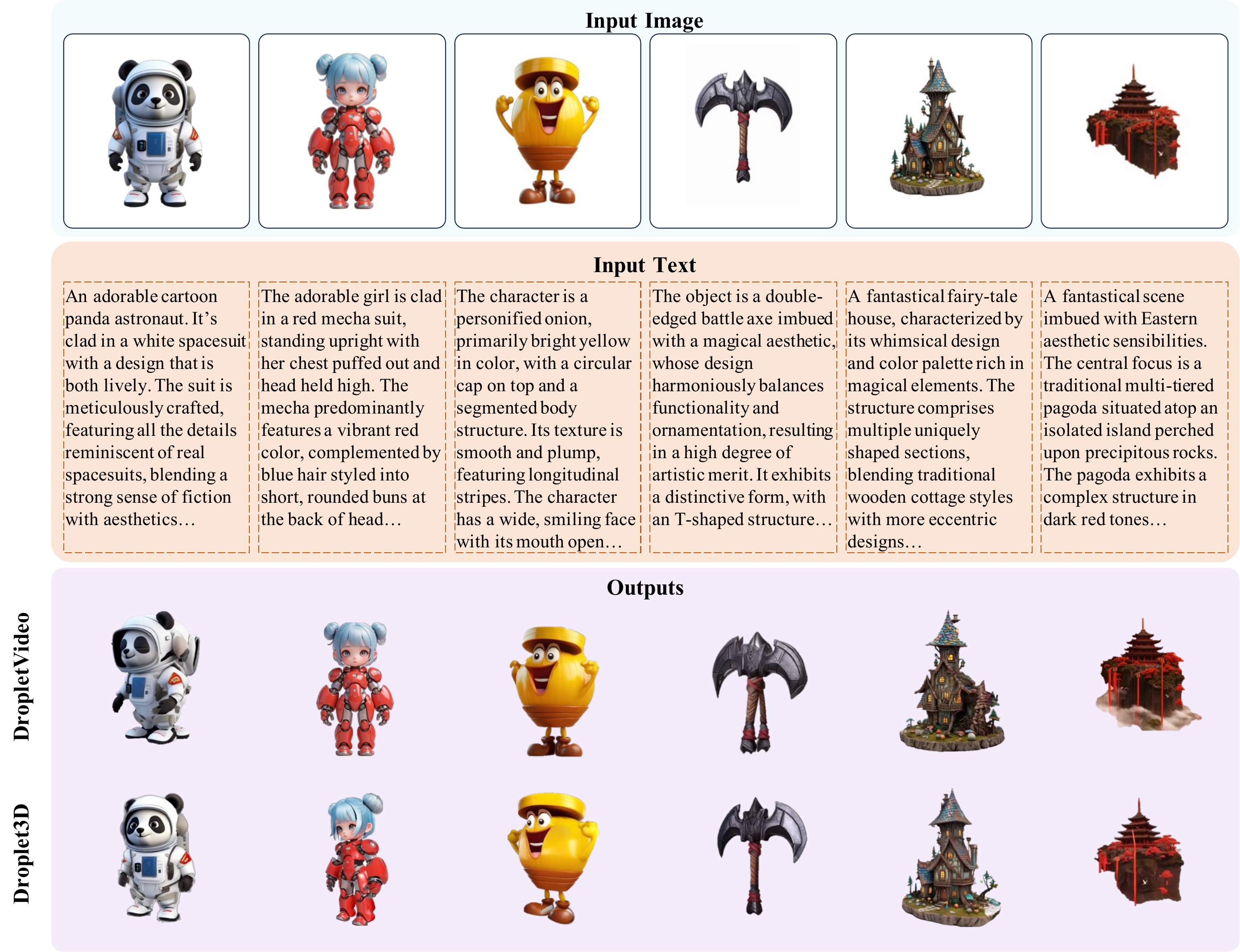}
\caption{\textbf{Comparison between \ourproj{} and its predecessor, DropletVideo.} The performance enhancement of \ourproj{} over DropletVideo in terms of spatial consistency underscores the critical role of continued training on the \ourdataset{} dataset.}
\label{fig:CompareDroplet}
\end{figure}

Considering that the video generation backbone network is a core module of \ourproj{}, we conduct quantitative comparisons using different backbone networks. 
These include DropletVideo\cite{zhang2025dropletvideo}, which serves as the genetic origin of \ourproj{}, and Cogvideox-Fun\cite{cogvideox-fun}, a model of comparable scale. 
Additionally, we compare against Wan2.1-I2V-14B and Step-Video-TI2V-30B, which are currently among the most state-of-the-art open-source video generation models.

\begin{table}[h]
\centering
\caption{\textbf{Comparison of \ourproj{} and the backbone it inherits from, \textit{DropletVideo}, in terms of generation performance.} \ourproj{} performs better.}
\label{tab:ExpBackbone1}
\small
\begin{tabular}{lcccccc}
\toprule
  & PSNR ($\uparrow$) & SSIM ($\uparrow$) & LPIPS ($\downarrow$) & MSE ($\downarrow$) & CLIP-S ($\uparrow$) \\
\midrule
\textbf{Droplet3D-5B} & 28.36 & 0.76 & 0.03 & 0.0017 & 0.866\\
DropletVideo-5B\cite{zhang2025dropletvideo} & 20.51 & 0.87 & 0.12 & 0.02 & 0.76\\
\bottomrule
\end{tabular}
\end{table}

\begin{figure}[h]
\centering
\includegraphics[width=1.0\linewidth]{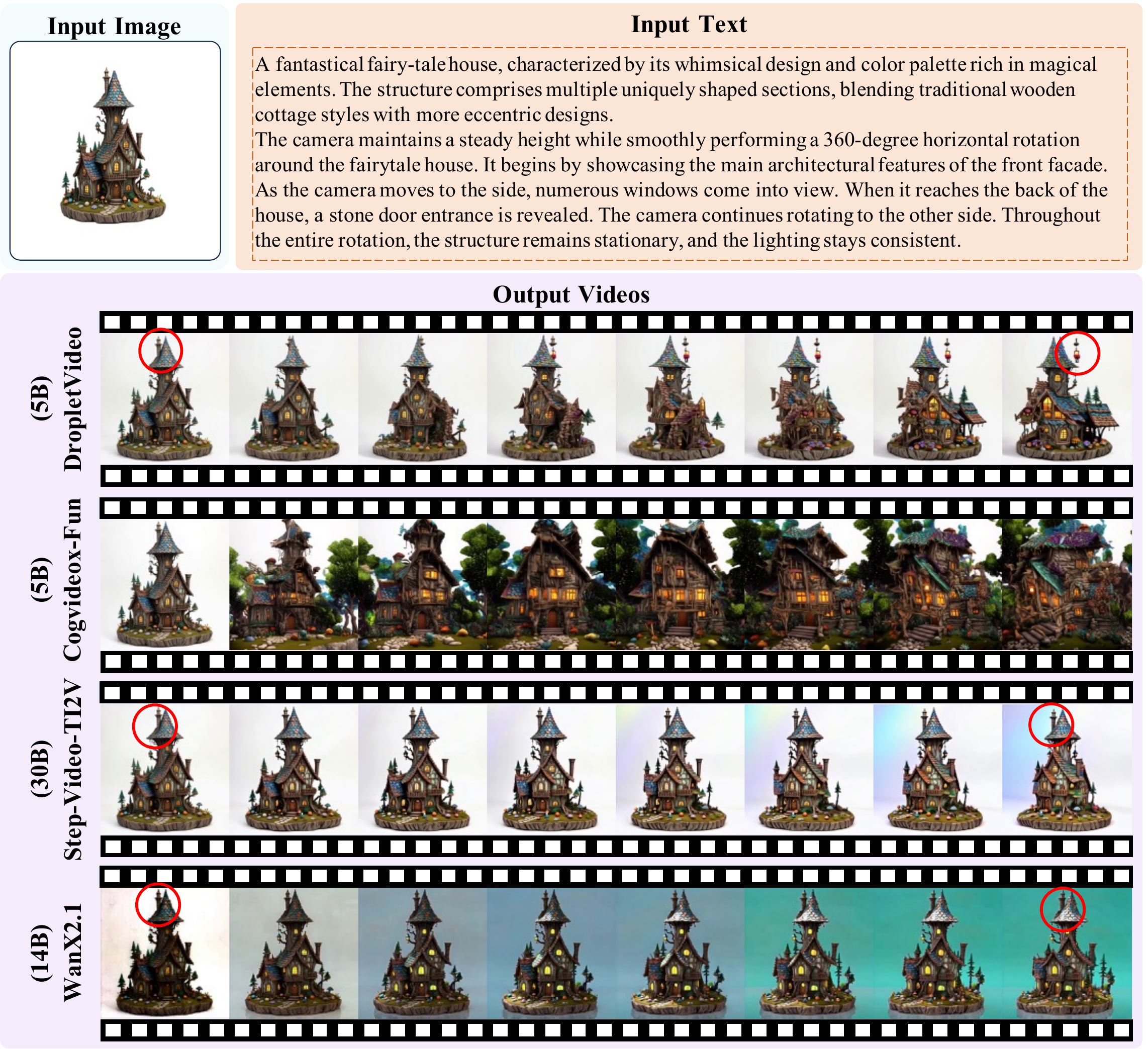}
\caption{\textbf{Comparison between DropletVideo and other video backbones.} DropletVideo demonstrates superior performance in circumnavigation shooting tasks. Using the chimney in \textbf{\textcolor{red}{red}} circle as an example, DropletVideo successfully generates surrounding-camera-motion videos, whereas Step and WanX only support minor rotational movements, and the similarly-sized Cogvideox-Fun lacks this capability entirely. Concurrently, DropletVideo also surpasses all competing models in preserving the original image’s style, such as its background color.}
\label{fig:CompareBackbone}
\end{figure}

\begin{table}[h]
\centering
\caption{\textbf{Comparison of different video generation backbone models}. We compared video models including \emph{DropletVideo}, in terms of generation performance.}
\label{tab:ExpBackbone2}
\small
\begin{tabular}{lccccccc}
\toprule
 & Params & PSNR ($\uparrow$) & SSIM ($\uparrow$) & LPIPS ($\downarrow$) & MSE ($\downarrow$) & CLIP-S ($\uparrow$) \\
\midrule
DropletVideo\cite{zhang2025dropletvideo} & 5B & 20.51 & 0.87 & 0.12 & 0.02 & 0.76\\
Cogvideox-Fun\cite{cogvideox-fun} & 5B & 15.16 & 0.50 & 0.21 & 0.09 & 0.68\\
Wan2.1-I2V & 14B & 12.97 & 0.89 & 0.19 & 0.067 & 0.67\\
Step-Video-TI2V & 30B & 15.32 & 0.79 & 0.30 & 0.08 & 0.64\\
\bottomrule
\end{tabular}
\end{table}

Tab. \ref{tab:ExpBackbone1} demonstrates the effectiveness of the proposed \ourdataset{} dataset, showing that models further trained with it exhibit improved generation consistency. 
It can be observed that while the original DropletVideo, which was not trained on \ourdataset{}, also demonstrates commendable performance across various metrics, it still falls short of the \ourproj{} proposed in this work.

Furthermore, as illustrated in Fig. \ref{fig:CompareDroplet}, DropletVideo demonstrates a notable capability in generating camera-rotating surround-view videos, with the synthesized content exhibiting high fidelity and geometric accuracy. 
This proficiency suggests that DropletVideo is well-suited to serve as a robust initialization for the weights of a backbone network. 
Nevertheless, the model exhibits certain ``shortcomings'' for 3D. Specifically, it tends to generate inconsistencies in dynamic objects, such as a panda depicted with its eyes closed in a static pose. 
However, this inherent limitation underscores the critical necessity for subsequent fine-tuning with the proposed \ourdataset{} to enhance the model’s performance in 3D generations.

In Fig. \ref{fig:CompareDroplet}, it can be observed that, in the provided examples, \ourproj{} is capable of generating superior 3D content.
A plausible explanation for this is that the generated surround-view images by \ourproj{} are both plausible and rich in detail, while maintaining high spatial consistency.
This endows our model with commendable foundational capabilities for 3D generation.

Additionally, Quantitative results shown in Tab. \ref{tab:ExpBackbone2} indicate that DropletVideo is indeed suitable for further training in 3D generation than other video generators. 
This may be related to its consideration of integral spatio-temporal consistency, as the large number of samples incorporating spatial consistency endows it with inherent 3D generation capabilities. 
It significantly outperforms other models of comparable scale, such as Cogvideox-Fun, and achieves performance comparable to or exceeding that of the other two larger and more advanced models, even without fine-tuning.
Tab. \ref{tab:ExpBackbone2} also demonstrates the video generation capabilities of DropletVideo-5B alongside Step-Video-TI2V-30B and Wan2.1-I2V-14B, currently the most popular open-source video generation models, despite their larger parameter counts. 
The results indicate that DropletVideo indeed possesses a higher capacity for maintaining 3D consistency compared to models of similar scale. 
While Step and WanX, due to their larger weights, also exhibit considerable ability, they may exhibit more pronounced object movements, such as the bear's head turning, the gold coin onion blinking or jumping, possibly due to the training samples being more inclined towards narrative changes. 
In contrast, CogVideoX-Fun-v1.5, with a similar parameter count to DropletVideo-5B, demonstrates poorer performance.
Visual examples are shown in Fig. \ref{fig:CompareBackbone}.
DropletVideo demonstrates a strong capability to generate multi-angle views of objects from input images, consistently achieving high-quality results for most cases.
Note that considering the trade-offs between spatial consistency, foundational generation capabilities, and model size, we selected DropletVideo as the backbone for subsequent training in the 3D domain.

\subsection{Potential and Applications in 3D Generation}

\subsubsection{Feasible Controllable-Creativity Driven by Language Prompts}

\begin{figure}[h]
\centering
\includegraphics[width=1.0\linewidth]{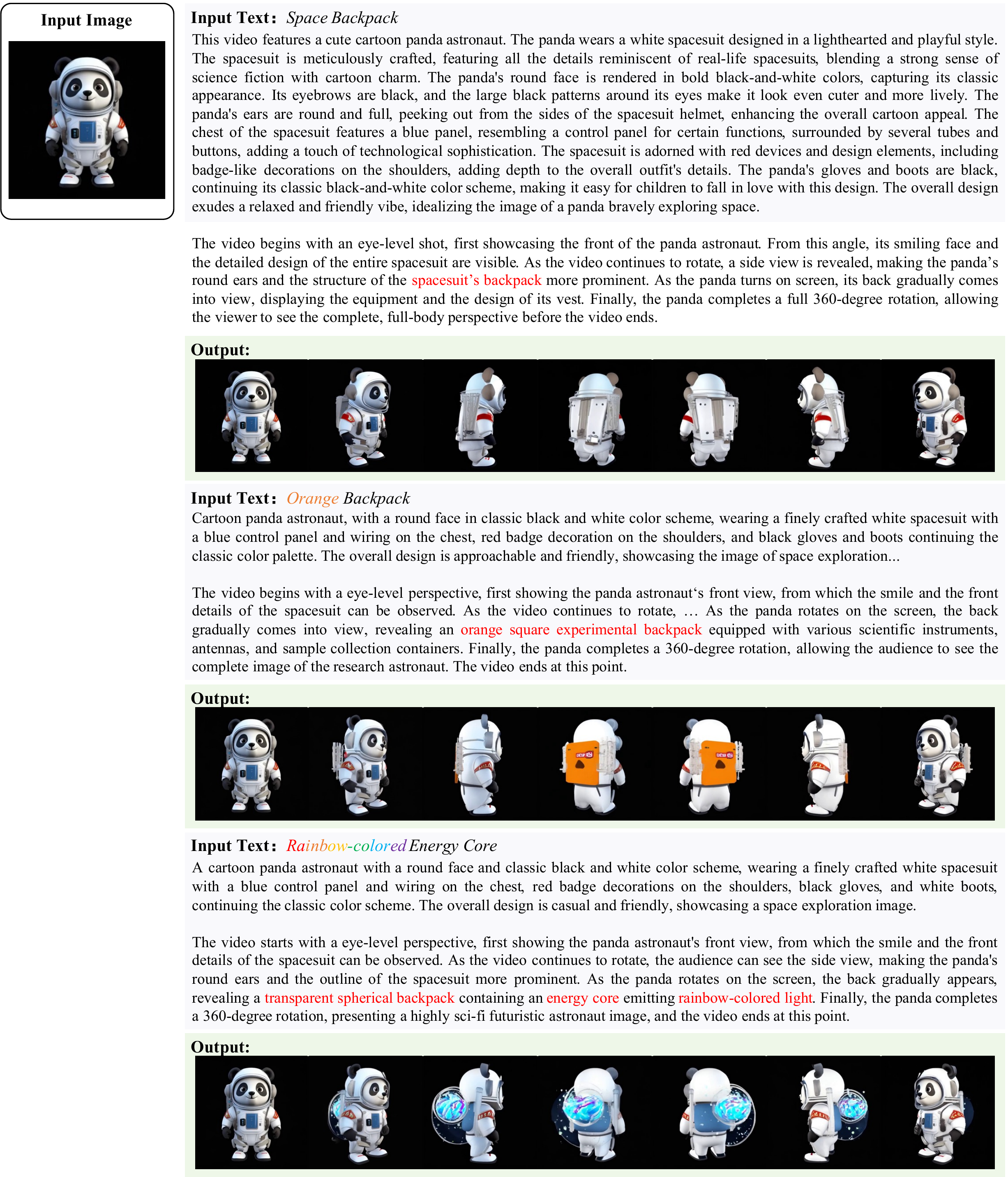}
\caption{\textbf{Controllable-Creativity based on the initial image of the Panda Astronaut case and different given texts.} The three rows in the figure respectively demonstrate the generation based on the given \textit{space backpack}, \textit{orange backpack}, and \textit{energy ball}.}
\label{fig:MultiPrompt_panda}
\end{figure}

\begin{figure}[h]
\centering
\includegraphics[width=1.0\linewidth]{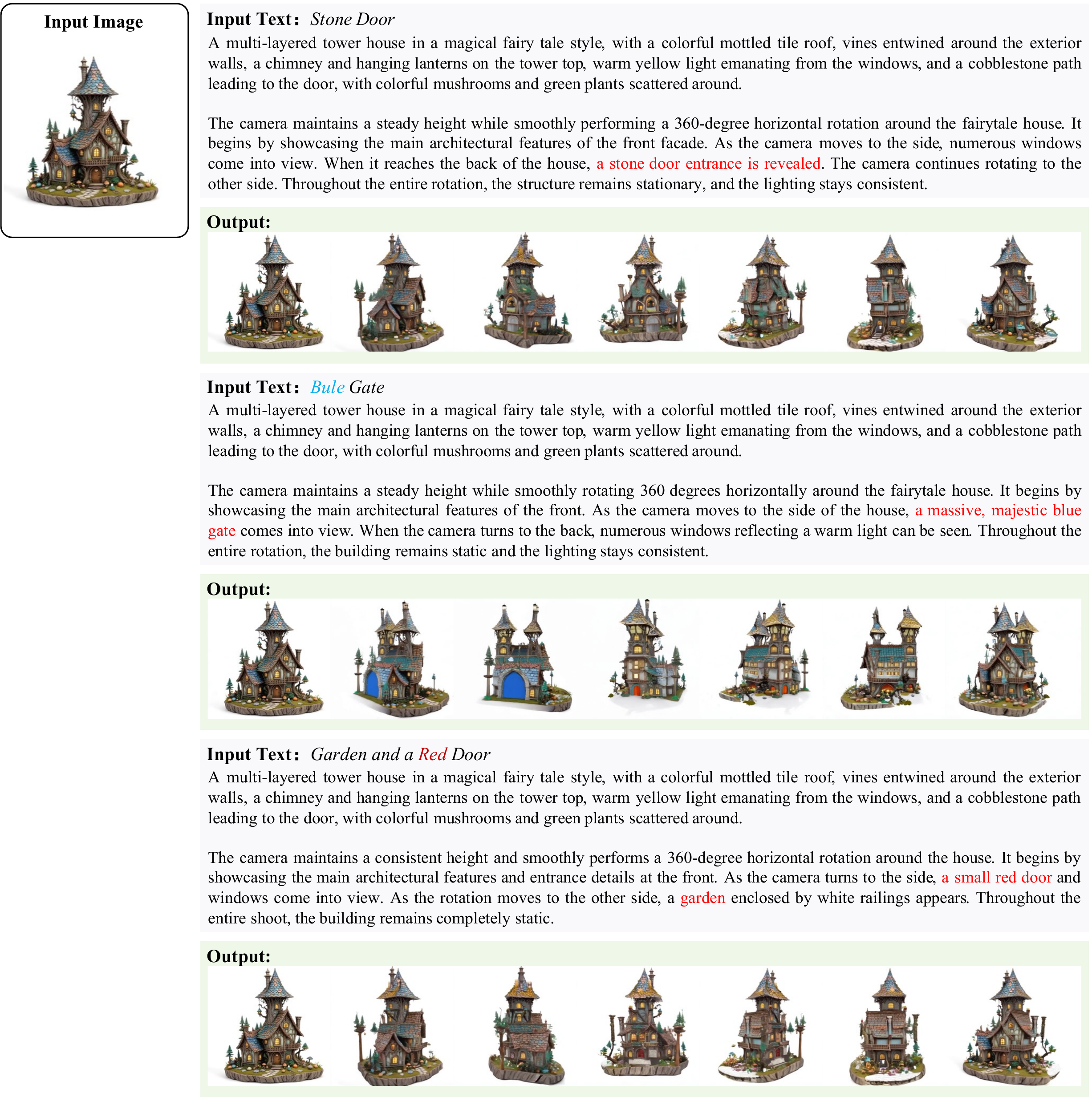}
\caption{\textbf{Controllable-Creativity based on the initial image of the Castle case and different given texts.} The three rows in the figure respectively demonstrate the generation based on the given \textit{stone door}, \textit{blue gate}, and \textit{a garden with a red door}.}
\label{fig:MultiPrompt_castle}
\end{figure}

\begin{figure}[h]
\centering
\includegraphics[width=1.0\linewidth]{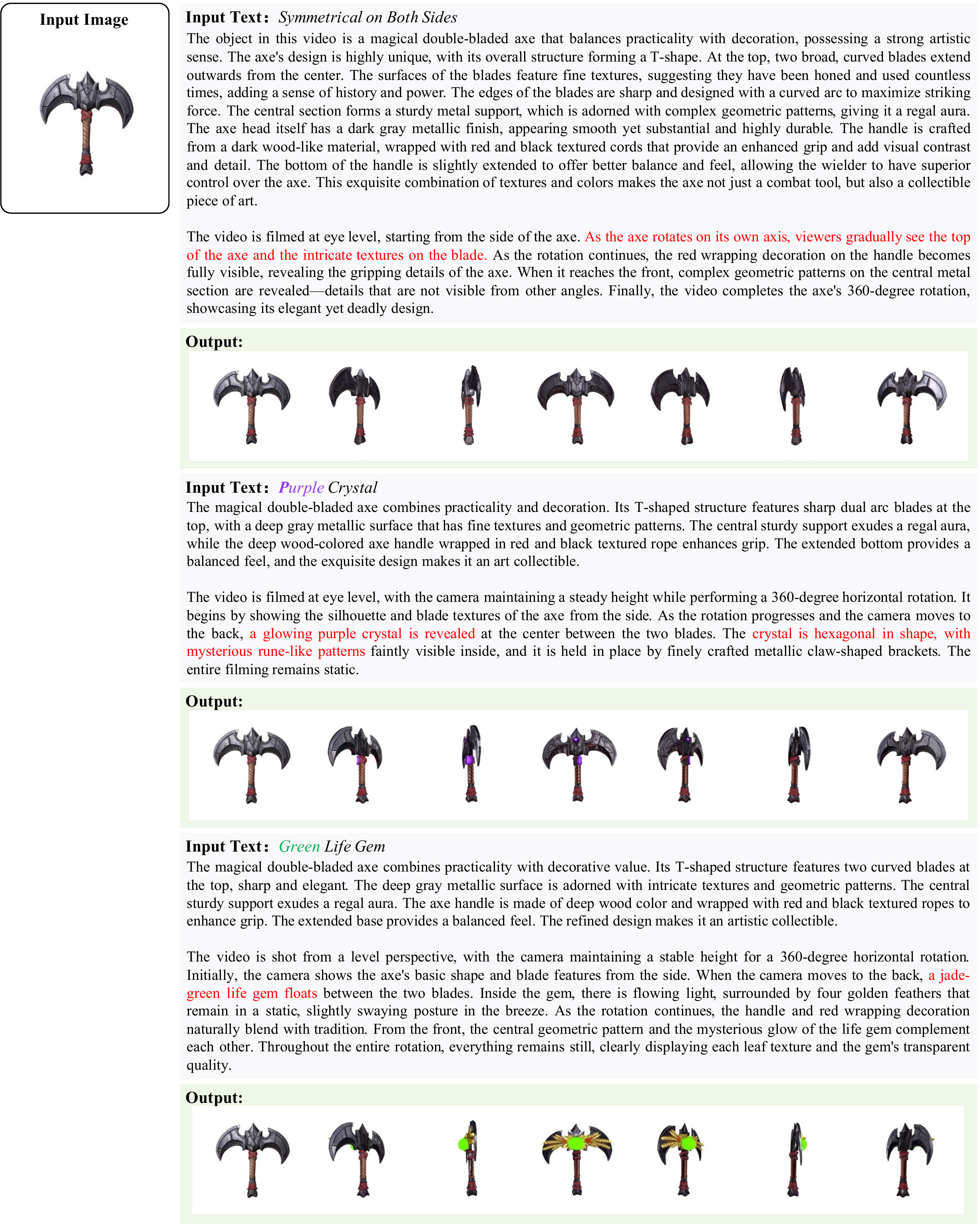}
\caption{\textbf{Controllable-Creativity based on the initial image of the Battle Axe case and different given texts.} The three rows in the figure respectively demonstrate the generation based on the given symmetrical side, a purple crystal, and a green life gem.}
\label{fig:MultiPrompt_axe}
\end{figure}

\begin{figure}[h]
\centering
\includegraphics[width=1.0\linewidth]{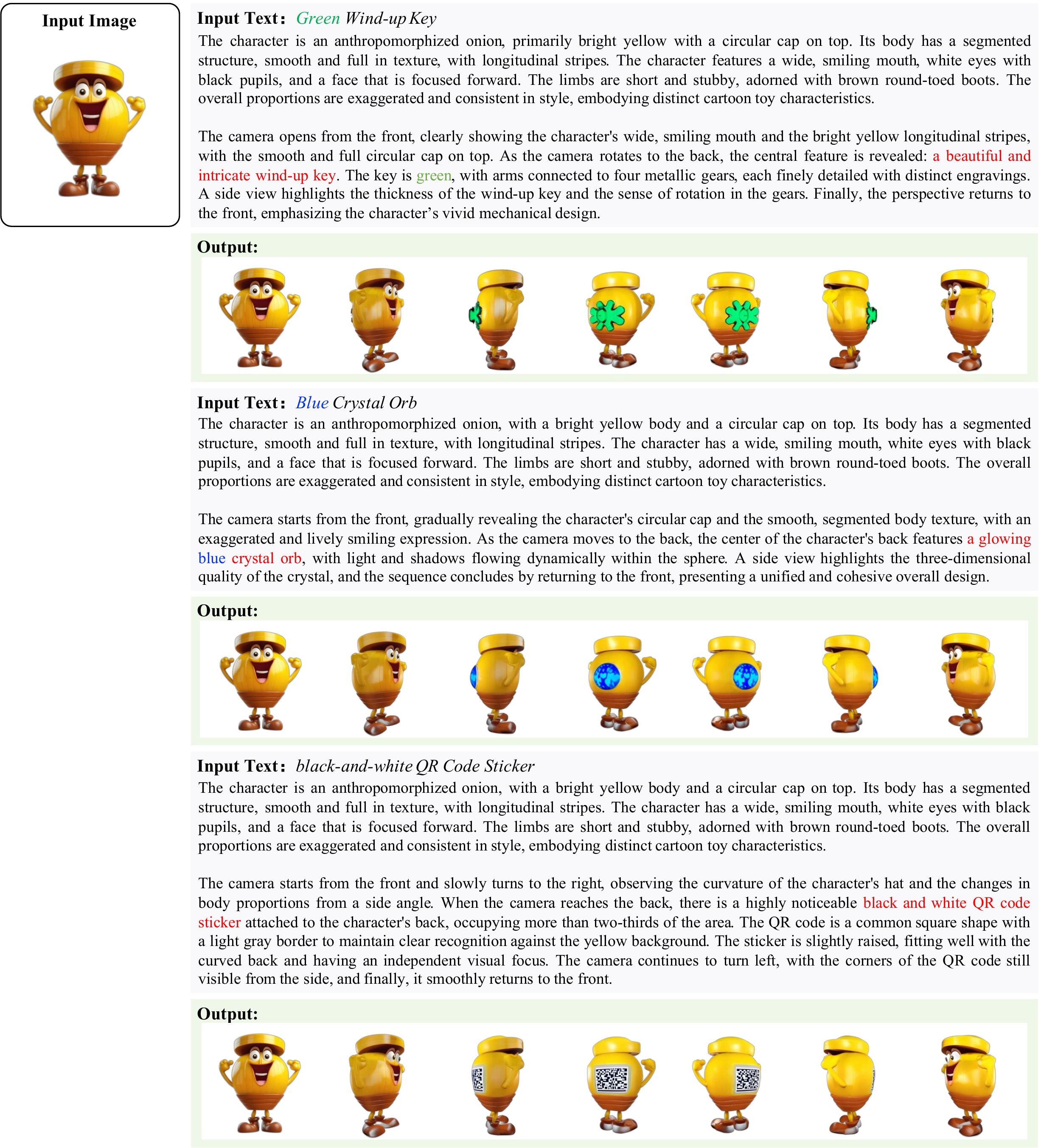}
\caption{\textbf{Controllable-Creativity based on the initial image of the Coin Onion case and different given texts.} The three rows in the figure respectively demonstrate the generation based on the given \textit{wind-up key}, \textit{crystal orb}, and \textit{QR code}.}
\label{fig:MultiPrompt_coin}
\end{figure} 

Current mainstream 3D generation models typically only support image or text input\cite{zhao2025hunyuan3d20,xiang2025structured}. 
However, \ourproj{} conditions on both an initial image and dense text, enabling it to support more targeted creative design, which is quite interesting. 
This is reflected in the variations of the generated assets when we employ AIGC images along with different textual descriptions.

Fig. \ref{fig:MultiPrompt_panda}–\ref{fig:MultiPrompt_coin} present four sets of examples. 
In each set, \ourproj{} generates different samples by utilizing the same AIGC-generated image as input, accompanied by distinct text guidance. 
As shown in Fig. \ref{fig:MultiPrompt_panda}, when different text descriptions are provided for the background of the \textit{panda astronaut}, the generated effects are both distinct and plausible. 
\ourproj{} can generate his \textit{space backpack}, \textit{orange backpack}, and \textit{a vibrant energy ball} that match the panda's physique and appearance based on its body shape and visual characteristics. 
Notably, in the example of generating \textit{energy ball}, \ourproj{} can produce objects that are either dazzling or semi-transparent. 
Compared to generating textured meshes, images or videos can more intuitively produce content with richer colors. 
This is also one of the factors we believe contributes to the feasibility of video-based generative models in 3D generation.
Similarly, as illustrated in Fig. \ref{fig:MultiPrompt_axe}, \ourproj{} can produce varying details behind the \textit{axe} depending on whether the text description specifies \textit{gems} or \textit{runes}. 
Besides, as shown in Fig. \ref{fig:MultiPrompt_coin}, when different decorations such as a \textit{wind-up key}, \textit{crystal orb}, or even a \textit{QR code} are input, the model is able to understand the semantic meaning of the text. 
Additionally, its color generation is also relatively accurate.
This capability for controllable 3D content generation stems from two main sources. 
On the one hand, our \ourproj{} inherently supports both image and text interfaces at the data level. 
On the other hand, it benefits from a richer prior learned from a vast volume of video media. 
First, the greater amount of spatial consistency data extracted from these videos helps to enhance the 3D generator’s ability to produce accurate content. 
Second, the more general semantic knowledge integrated from the video backbone model enables our model to better comprehend textual requirements. 
A prime example is the ``QR code'' generation task. 
When prompted to generate a ``QR code'', an object scarcely found in 3D model datasets, \ourproj{} can still accurately identify what it is and generate it correctly, drawing upon its knowledge derived from videos.

These experimental results demonstrate the potential of our proposed method in the field of 3D asset generation. 
The ability to create based on both images and expressions is one of the attributes we find particularly compelling, and it also validates the feasibility of leveraging video to facilitate 3D creation.
Unlike generation algorithms that support only a single modality, either image or text, simultaneously referencing image priors and supplementing with textual descriptions also holds practical utility for 3D content generation.

\subsubsection{3D Lifting from Stylized Inputs}

\begin{figure}[htbp]
\centering
\includegraphics[width=1.0\linewidth]{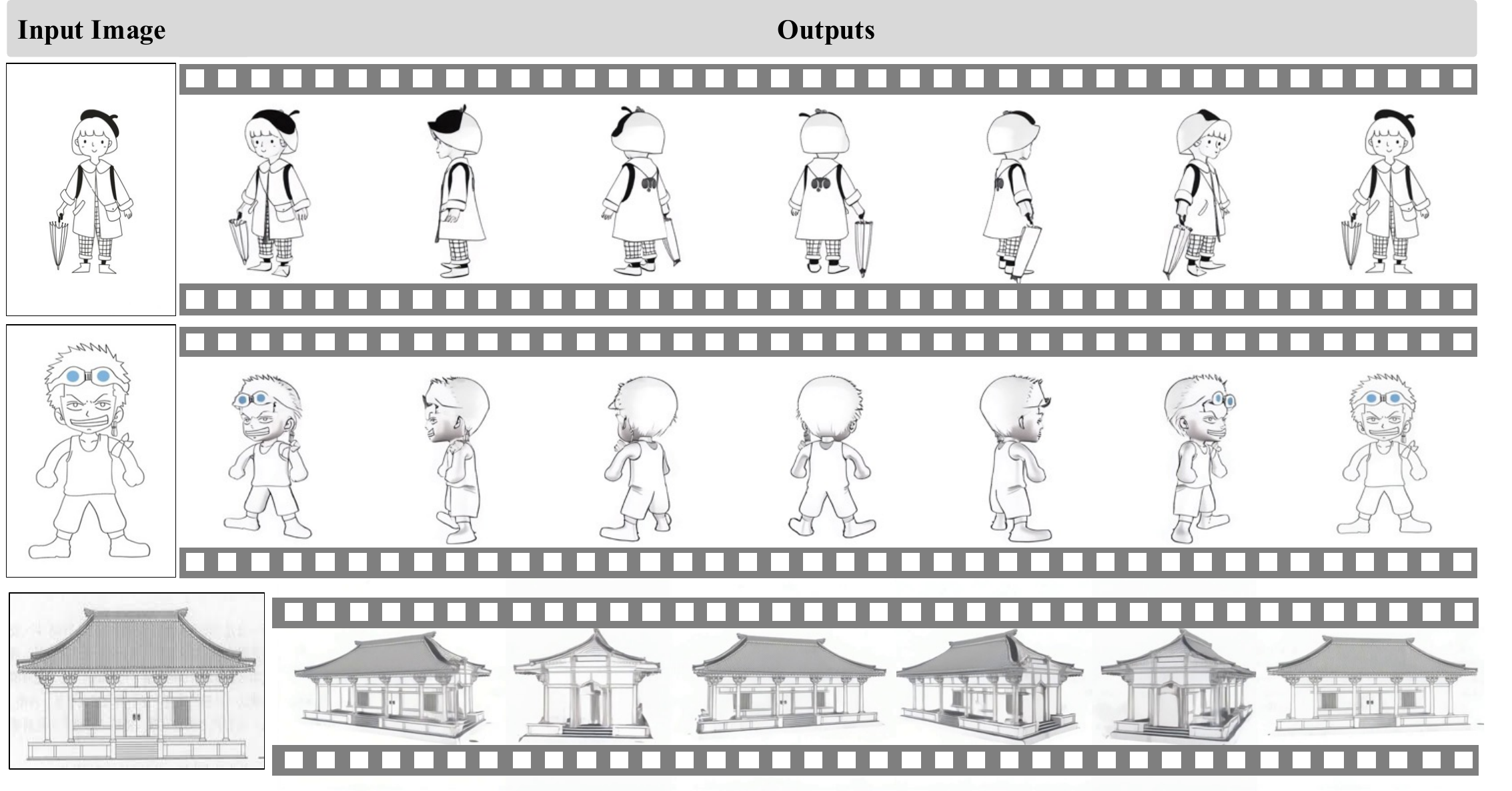}
\caption{\textbf{\ourproj{} demonstrates its lifting capability on 2D sketch paintings.} Our model can transform objects such as characters and buildings into three-dimensional representations based on simple line drawings. From up to bottom: a sketch of a girl dressed in a minimalist style, a Japanese anime character, and a traditional Chinese architectural structure.}
\label{fig:Lifting_sketch}
\end{figure}

\begin{figure}[htbp]
\centering
\includegraphics[width=1.0\linewidth]{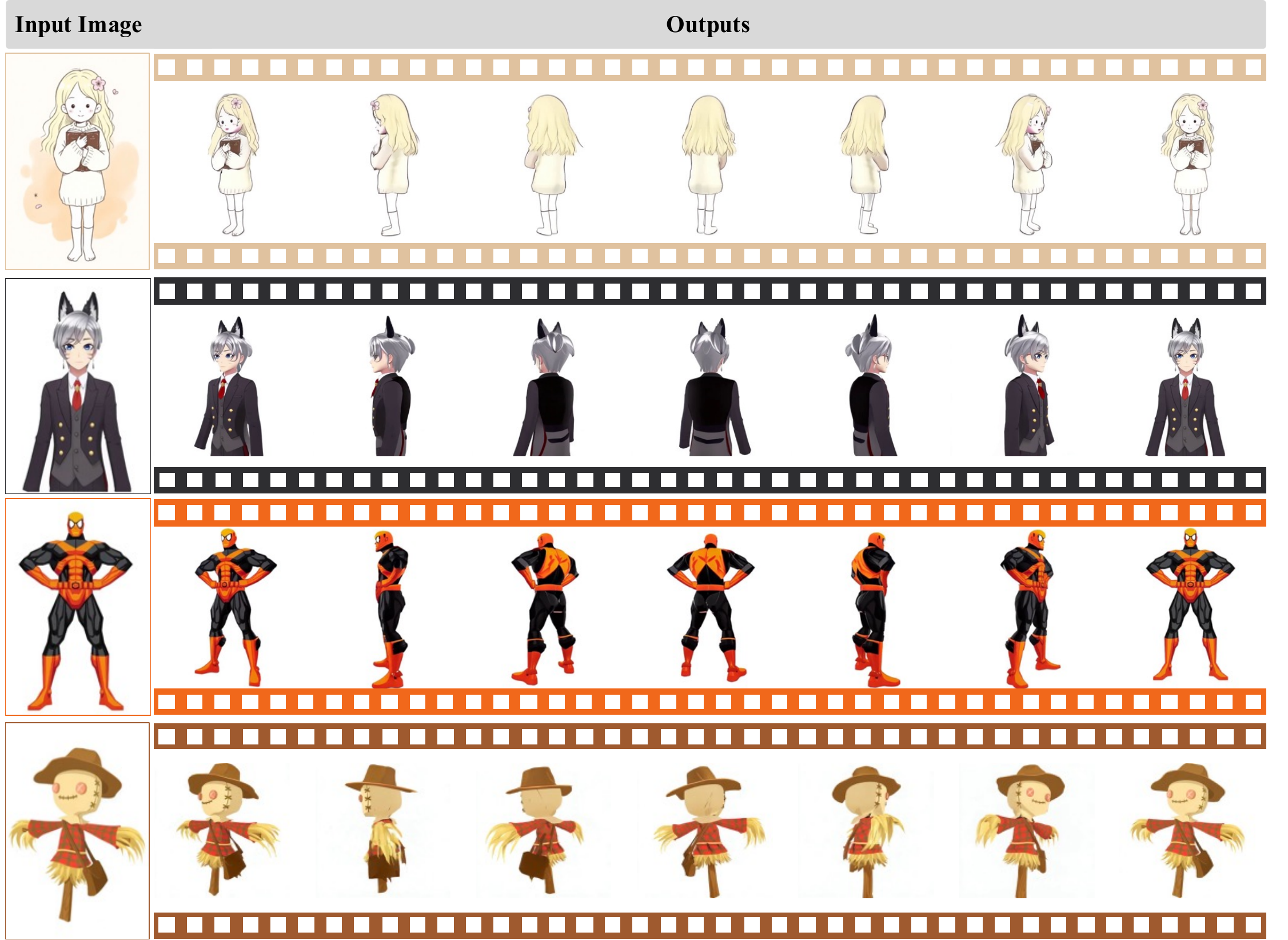}
\caption{\textbf{\ourproj{} demonstrates its lifting capability on 2D images styled as comics paintings.} Our model can elevate objects within a given 2D comic into 3D, thereby achieving a cross-dimensional effect. From top to bottom: a gentle girl in a comic style, a male student with animal ears, an American comic-style superhero, and a scarecrow in the style of Hayao Miyazaki.}
\label{fig:Lifting_comics}
\end{figure}

\ourproj{} exhibits a relatively robust capability in lifting 2D content into 3D. 
Although the training data for \ourdataset{} is entirely based on renderings from Objaverse-XL, which consists solely of object-level 3D files, \ourproj{} still demonstrates a certain degree of robustness, particularly when the input images differ significantly from the training data distribution, such as stylized AIGC images like comics or sketches. 
We believe this capability may stem from its prior video training, which has endowed it with extensive general knowledge, making its 3D generation more versatile. 
This observation is particularly interesting, as it validates, to some extent, our hypothesis that video can facilitate 3D generation. 
Figs. \ref{fig:Lifting_sketch} and \ref{fig:Lifting_comics} demonstrate the lifting capacity for sketches and comic-style images, respectively.

On the left of each figure are the stylized image inputs, while the right shows the surrounding perspective segments generated by our model based on it. 
For example, as shown in the first row of Fig. \ref{fig:Lifting_comics}, our model can identify the body shape and posture of the comic girl and smoothly transform it into a 3D form without altering its style.
We believe that this capability to support stylized images contributes to the generation of 3D assets. For instance, as shown in Fig. \ref{fig:Lifting_sketch}, 
it holds significant potential to form a well-developed 3D object based on just a few simple strokes from a designer or illustrator, whether it be a character (1st and 2nd rows) or an architecture (3rd row).

\subsubsection{Textured Mesh Synthesis}

\begin{figure}[htbp]
\centering
\includegraphics[width=1.0\linewidth]{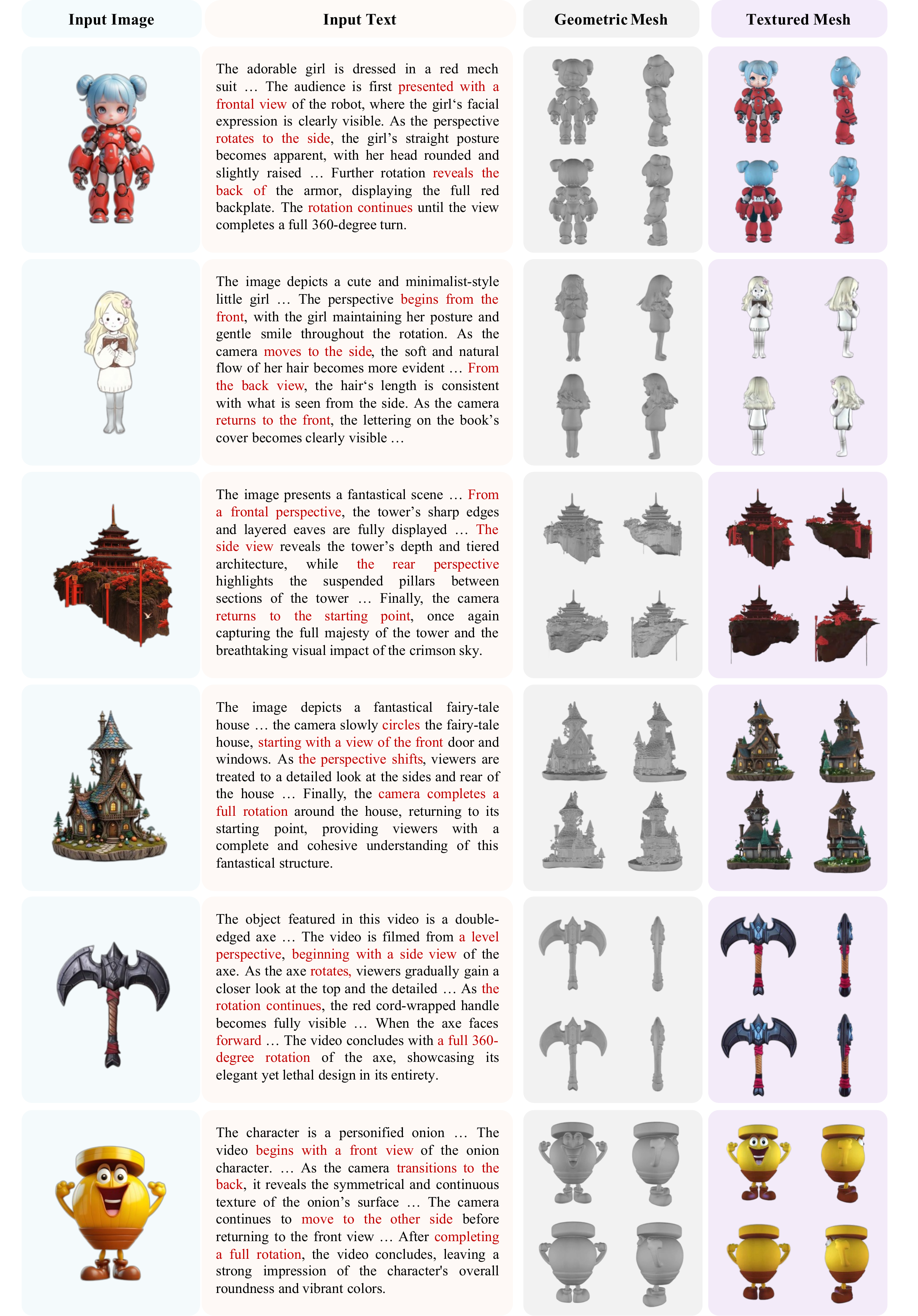}
\caption{\textbf{Mesh reconstruction based on content generated by \ourproj{}.} Note that the models used for mesh generation and texture mapping originate from the open-source Hunyuan3D-2.0 and Hunyuan3D-2.1, respectively.}
\label{fig:App_mesh}
\end{figure}

We tested the potential of \ourproj{} to create practical and usable 3D assets.
Fig. \ref{fig:App_mesh} illustrates the results of generating textured meshes using \ourproj{}. 
We employed Hunyuan3D-2\cite{zhao2025hunyuan3d20} as the synthesis tool. 
Specifically, we utilized Hunyuan3D-2.0 DiT model for the geometric mesh synthetic and employed Hunyuan3D-2.1 painting model to generate textured meshes. 
We applied this algorithm to convert the multi-view images from \ourproj{} into an industrially applicable	 representation.
Compared to other mesh generation models that directly utilize single images, our pipeline enables the creation of a more diverse range of appearances based on long or short text descriptions.

\subsubsection{3D Gaussian Splatting Points Synthesis}

\begin{figure}[htbp]
\centering
\includegraphics[width=1.0\linewidth]{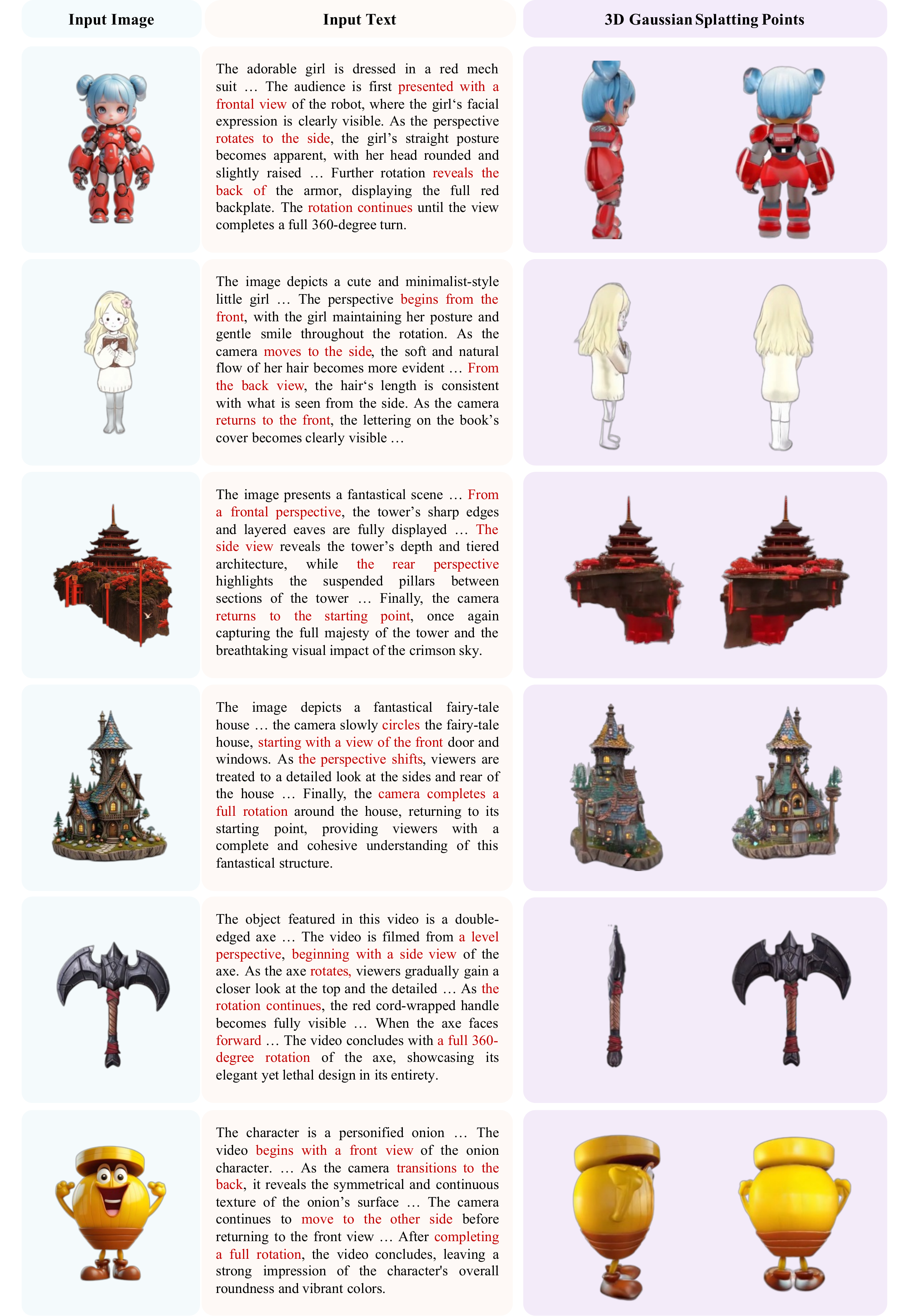}
\caption{\textbf{3D Gaussian splatting reconstruction based on content generated by \ourproj{}.} We employed the native optimization-based approach for this implementation. }
\label{fig:App_3dgs}
\end{figure}

In addition, we also tested the potential to create 3D Gaussian splatting modality contents.
Fig. \ref{fig:App_3dgs} presents examples of 3D Gaussian splatting generated using \ourproj{}. 
We used the native optimization-based approach\cite{kerbl20233d} for this implementation. 
As can be observed, the re-rendered images maintain high quality, demonstrating the effectiveness of our proposed method. 
The generated surrounding viewpoints exhibit strong spatial consistency. 
Considering that we only employ the most basic reconstruction algorithm without any post-processing steps such as denoising, completion or camera pose estimation, we argue that a plausible reason is that our model can produce dense viewpoints, with the surrounding motion poses being relatively uniform and accurate.

\subsubsection{Scene-level 3D Content Generation}

\begin{figure}[htbp]
\centering
\includegraphics[width=1.0\linewidth]{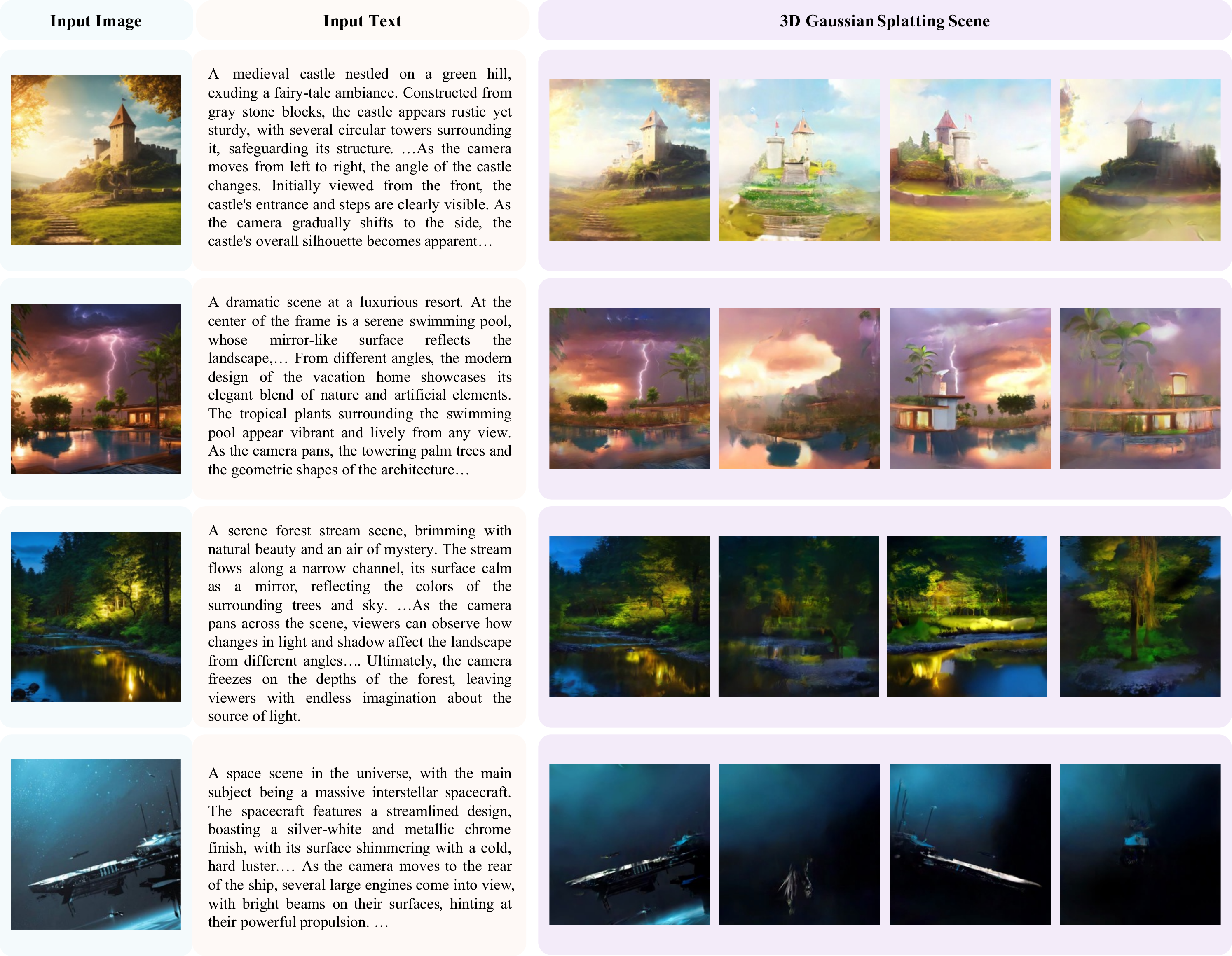}
\caption{\textbf{Gaussian splatting edition based on scene-level content generated by \ourproj{}.} Compared to existing mainstream object-level 3D generation methods, \ourproj{} not only generates individual objects but also demonstrates significant potential for scene-level content generation.}
\label{fig:App_3dgsscene}
\end{figure}

Additionally, what excites us most is that, as illustrated in Fig. \ref{fig:App_3dgsscene}, our model can perform lifting on images with scene-level styles, which validates that our technical approach is fundamentally distinct from other currently popular native 3D generation methods, which are typically \textbf{object-level}. 
In contrast, \ourproj{} has the potential to convert \textbf{scene-level} content into 3D.
Fig. \ref{fig:App_3dgsscene} sequentially demonstrates scene-level 3D generation for a manor, an island with lightning, a tranquil riverside at night, and a scene deep within a space station.
Similar to a cinematic freeze-frame effect, our model can automatically lift these scenes to 3D, thereby reducing labor costs.
From this perspective, it can also be regarded as another form of 3D asset creation.
However, what is even more intriguing is that the training set, \ourdataset{}, contains no scene-level samples. 
Therefore, this capability can be considered entirely inherited from its ancestral source, the DropletVideo video generation model.

Fig. \ref{fig:App_3dgsscene} demonstrates the reconstruction of scene-level results using Gaussian splatting. 
It is interesting that we can further develop more applications based on these generated contents. 
For example, users can use these scene contents and combine it with the object-level point set. 
In terms of implementation, we can embed objects into the scene by controlling the scale, displacement, and rotation of both point sets. 
This intriguing application is intrinsically linked to \ourproj{}'s potential for generating scene-level content.

\section{Conclusion}

In this paper, we propose a technical approach for 3D generation based on video backbone models, aiming to leverage the vast repository of video media materials to enhance the general knowledge capabilities of 3D generation, thereby facilitating the creation of 3D assets. 
On one hand, we introduce a large-scale 3D training dataset, termed \ourdataset{}, comprising 4 million 3D models, accompanied by high-quality rendered surround-view videos and viewpoint-level textual descriptions. 
On the other hand, we present the technical framework of \ourproj{} and achieve the training of 3D model weights using the DropletVideo, a video generation backbone network with spatio-temporal consistency. 
Concurrently, we design a text rewriting module and an image canonical viewpoint alignment module to refine the pipeline. 
Finally, we substantiate the value of our proposed approach through extensive visualization experiments. 
Furthermore, comprehensive quantitative experiments demonstrate its foundational capabilities and effectiveness. 
We hope that the ideas presented in this paper will contribute to the advancement of the 3D generation algorithm community, and the relevant dataset, code, and model weights associated with this work have been made open-source.

\bibliographystyle{plain}
\bibliography{sample_3d}

\end{document}